\documentclass[sigconf]{acmart}

\AtBeginDocument{%
  }

\copyrightyear{2025}
\acmYear{2025}
\setcopyright{acmlicensed}\acmConference[MM '25]{Proceedings of the 33rd ACM International Conference on Multimedia}{October 27--31, 2025}{Dublin, Ireland}
\acmBooktitle{Proceedings of the 33rd ACM International Conference on Multimedia (MM '25), October 27--31, 2025, Dublin, Ireland}
\acmDOI{10.1145/3746027.3754938}
\acmISBN{979-8-4007-2035-2/2025/10}

\usepackage{multirow} 

\begin{document}

\title{AStF: Motion Style Transfer via Adaptive Statistics Fusor}


\author{Hanmo Chen}
\authornote{Both authors contributed equally to this research.}
\email{hmc@stu.xidian.edu.cn}
\orcid{0009-0005-8849-1936}
\affiliation{%
  \institution{Hangzhou Institute of Technology, Xidian University}
  \city{Hangzhou}
  \state{Zhejiang}
  \country{China}
}

\author{Chenghao Xu}
\authornotemark[1] 
\email{chx@stu.xidian.edu.cn}
\orcid{0000-0001-5888-0504}
\affiliation{%
  \institution{Xidian University}
  \city{Xi'an}
  \state{Shaanxi}
  \country{China}
}

\author{Jiexi Yan}
\authornote{Corresponding author}
\email{yanjiexi@xidian.edu.cn}
\orcid{0000-0002-2544-3057}
\affiliation{%
  \institution{Xidian University}
  \city{Xi'an}
  \state{Shaanxi}
  \country{China}
}

\author{Cheng Deng}
\authornotemark[2] 
\email{chdeng@mail.xidian.edu.cn}
\orcid{0000-0003-2620-3247}
\affiliation{%
 \institution{Xidian University}
  \city{Xi'an}
  \state{Shaanxi}
  \country{China}
 }



\begin{abstract}
Human motion style transfer allows characters to appear less rigidity and more realism with specific style. Traditional arbitrary image style transfer typically process mean and variance which is proved effective. Meanwhile, similar methods have been adapted for motion style transfer. However, due to the fundamental differences between images and motion, relying on mean and variance is insufficient to fully capture the complex dynamic patterns and spatiotemporal coherence properties of motion data. Building upon this, our key insight is to bring two more coefficient, skewness and kurtosis, into the analysis of motion style. Specifically, we propose a novel Adaptive Statistics Fusor (AStF) which consists of Style Disentanglement Module (SDM) and High-Order Multi-Statistics Attention (HOS-Attn). We trained our AStF in conjunction with a Motion Consistency Regularization (MCR) discriminator. Experimental results show that, by providing a more comprehensive model of the spatiotemporal statistical patterns inherent in dynamic styles, our proposed AStF shows proficiency superiority in motion style transfers over state-of-the-arts. Our code and model are available at \url{https://github.com/CHMimilanlan/AStF}.

\end{abstract}

\begin{teaserfigure}
  \includegraphics[width=\textwidth]{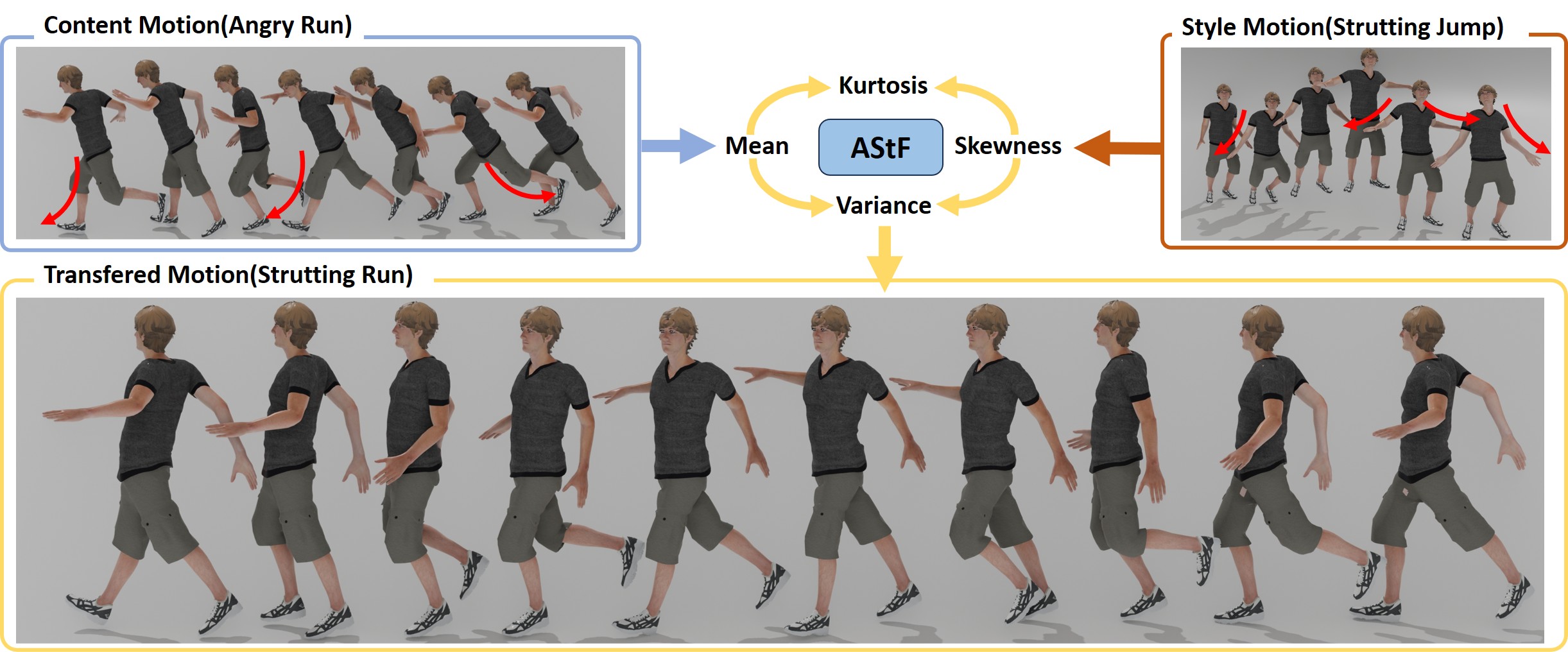}
  \caption{Motion style transfer via Adaptive Statistics Fusor(AStF). \textnormal{The left red solid curved represents the content we desire to preserve, while the right red solid curved arrow indicates the action with style characteristics, which is the style we aim to transfer.}}
  \label{fig:teaser}
\end{teaserfigure}


\maketitle

\section{Introduction}

Recent years have witnessed growing attention on computer-generated human animation ~\cite{stark2024animation, mourot2022survey, bi2025llavasteeringvisualinstruction, bi2025cot}, owing to its critical role in bridging the gap between virtual character behaviors and human perceptual realism ~\cite{xu2024rethinking}. This technology finds extensive applications in virtual reality system, video game, and anime. Within this domain, human motion constitutes a fundamental component whose quality directly determines animation verisimilitude ~\cite{bian2024motioncraft, zeng2024light, mcallister2025behavioural, bi2025prismselfpruningintrinsicselection}. In the real world, human motions express multiple characteristics including emotion, physiological conditions, age et al., which collectively form motion style descriptors. This understanding has propelled motion style transfer into a forefront research challenge ~\cite{zhu2023human, loi2023machine, xu2025keep}. The primary objective of motion style transfer is to seamlessly transfer stylistic traits from a stylized motion sequence to another content motion sequence. 


The primary challenge in motion style transfer lies in effectively disentangling and integrating motion style while preserving the physical plausibility of the content~\cite{chang2022unifying, tao2022style, jiang2025packdit, xu2024llm}. Early approaches~\cite{holden2016deep, holden2017fast} employed feedforward neural networks to disentangle motion styles, enabling direct style transfer. However, these methods often lacked diversity in the generated styles. More recently, inspired by the success of image style transfer techniques such as AdaIN~\cite{huang2017arbitrary}, AdaAttn~\cite{liu2021adaattn}, and StyA2K~\cite{zhu2023all}, contemporary motion style transfer methods~\cite{kim2024most, guo2024generative, jang2022motion} have widely adopted the AdaIN paradigm. This approach adaptively modulates the normalized content features using the mean and standard deviation of the style features, aligning the statistical distribution of the content features with that of the style, thereby achieving effective style transfer.


Despite the significant progress achieved, the effectiveness of existing AdaIN-based methods remains limited when applied to motion style transfer due to fundamental differences between static images and dynamic motions. In image style transfer, style can be mostly regarded as an overall mixture of colors and textures, essentially analogous to a photographic filter. In most image style transfer cases, stylization intensity contributes equally to all region of an image without noticeable asymmetry. It is uncommon for a small region to be heavily stylized while another remains unaffected. Therefore, utilizing mean and variance is sufficient to capture perceptually relevant style information. However, motion styles inherently involve complex spatiotemporal dynamics and asymmetric variations across multiple dimensions ~\cite{fang2024cigtime, athanasiou2024motionfix, yang2024unimumo}, including joint position, velocity, and acceleration, as well as trajectory patterns. For example, angry motion styles often manifest as increased limb acceleration, whereas styles associated with elderly individuals exhibit slower movement velocities. Even within the same motion content, such as a punch, different joints may exhibit distinct dynamic behaviors—e.g., rapid arm acceleration while leg movement remains minimal. Moreover, a single joint can display diverse temporal behaviors across different time steps, further illustrating the asymmetric and temporally variant nature of motion ~\cite{li2024unimotion, basset2024smear}.

\begin{figure}[t]
  \centering
  \includegraphics[width=\linewidth]{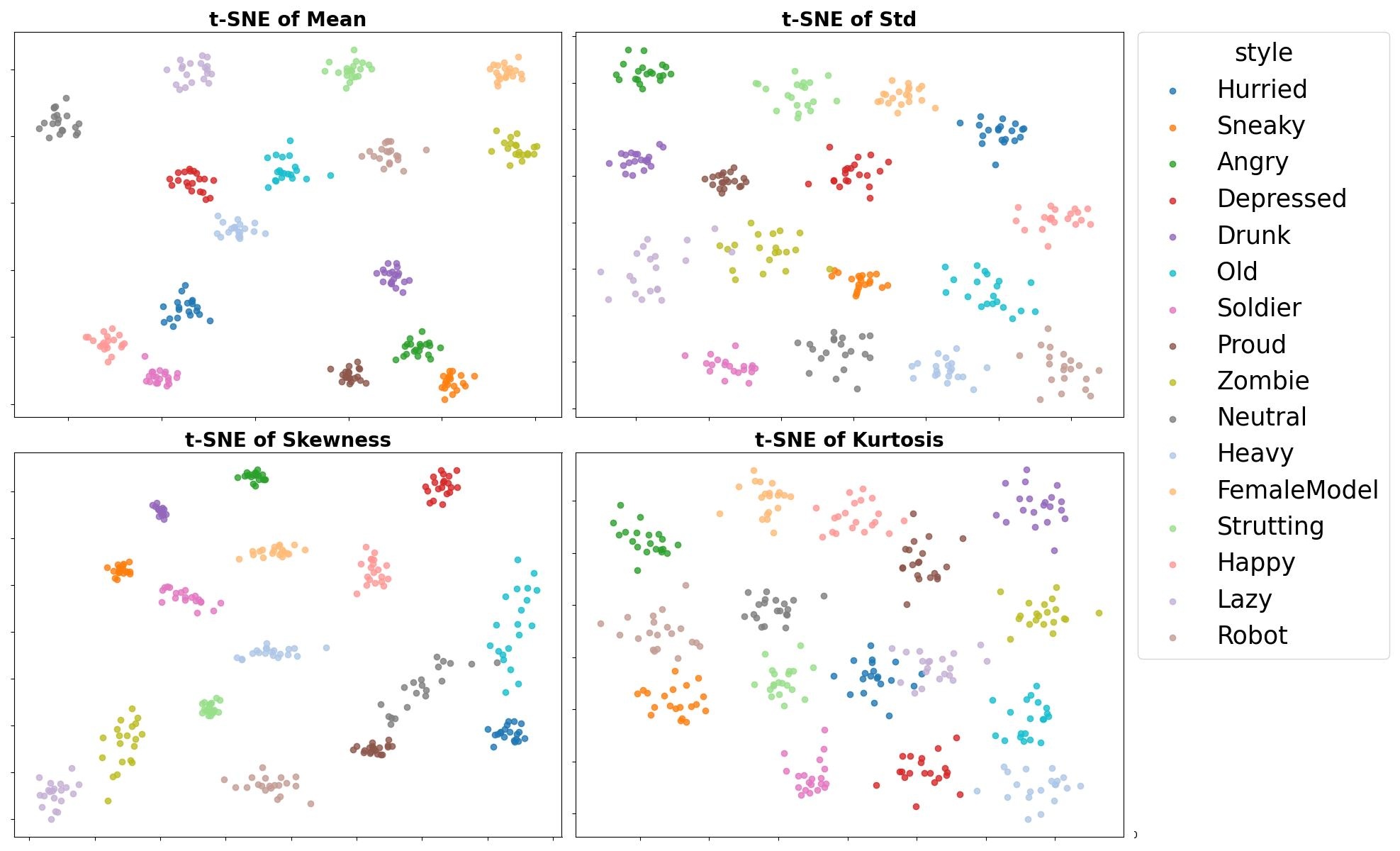}
  \caption{Statistics distribution differences across styles by t-SNE on BFA dataset ~\cite{aberman2020unpaired}.}
\label{fig:tsne}
\end{figure}

These complex motion dynamics suggest that effective motion style transfer necessitates modeling beyond global first- and second-order statistics, with particular emphasis on higher-order and joint-specific temporal characteristics. While prior approaches relying solely on mean and variance offer a foundation for style disentanglement, they are inadequate for capturing the nuanced dynamics and asymmetries inherent in motion data. Building on this observation, we explore the incorporation of higher-order statistical descriptors. To validate our hypothesis, we visualize the distribution of various motion styles under different statistical metrics using t-SNE~\cite{van2008visualizing}, as shown in Figure~\ref{fig:tsne}. The resulting scatter plot reveals substantial divergence in the distribution patterns across styles when higher-order statistics are considered. These findings suggest that metrics such as skewness and kurtosis may provide a more expressive and discriminative characterization of motion styles, motivating their integration into the modeling framework.

In this paper, inspired by Zhang et al. ~\cite{zhang2025hsi}, we introduce skewness and kurtosis as higher-order dynamic statistics and propose a novel Adaptive Statistics Fusor (AStF), comprising a Style Disentanglement Module (SDM) and a High-Order Multi-Statistics Attention mechanism (HOS-Attn). Within the AStF framework, skewness captures the asymmetry in motion sequence distributions, while kurtosis reflects the dynamic intensity and variability of motion styles. Together with mean and variance, these four statistical measures constitute a comprehensive spatiotemporal representation that effectively disentangles the latent characteristics of motion styles. Following the extraction of these statistical features, the HOS-Attn module is designed to adaptively inject the disentangled style statistics into the content motion through spatiotemporal-aware weighting, ensuring both effective style integration and preservation of temporal coherence.

Additionally, contrastive learning is often used in motion style transfer, where discriminator significantly influences the generator. Existing methods \cite{kim2024most, jang2022motion, park2021diverse} have not focused much on the configuration of the discriminator, as they simply input generated motion and style motion into the discriminator separately to compute adversarial loss, leading to a lack of correlation between generated motion and style motion within the discriminator. Therefore, inspired by Ko et al. ~\cite{ko2022self}, we propose the Motion Consistency Regularization (MCR) discriminator, which utilizes a correlation loss to guide the discriminator in capturing style consistency between the generated motion and the reference style motion. 

To summarize, our contributions are listed as follows:

\begin{itemize}
	\item  We designed AStF by incorporating SDM and HOS-Attn. SDM effectively disentangles four statistics to capture the global, asymmetry, and dynamic features of motion, while HOS-Attn integrates the disentangled style statistics into the content motion, maintaining spatiotemporal coherence.
	\item  We propose MCR discriminator which significantly enhances the style consistency of generated results in motion style transfer by increasing the feature similarity between style motion and generated motion, as well as the internal consistency within the style motion, in the discriminator.
        \item Our proposed AStF shows superiority performance in motion style transfer compared to existing methods.
\end{itemize}

\section{Related Works}

\subsection{Image Style Transfer}

Since stylized images possess certain artistic significance, style transfer is first applied to the image. Early research on image style transfer mainly relied on hand-craft features until Gatys et al. ~\cite{gatys2016image} pioneered transfer style from one image to another using a convolution neural network. To improve efficiency, Johnson et al.~\cite{johnson2016perceptual} combine the benefits of feedforward networks and perceptual loss functions, achieving faster and better transferring. Alternatively, CycleGAN (Zhu et al., 2017), transfer styles between images by utilizing Generative Adversarial Networks (GANs) with local similarity. Ulyanov et al.~\cite{ulyanov2016instance} enhanced stylization quality and better detail preservation by introducing Instance Normalization (IN). Building on IN, Huang et al.~\cite{huang2017arbitrary} introduced Adaptive Instance Normalization (AdaIN), which is a a significant advancement that adjusts the mean and variance of content features to match those of an arbitrary style image. However, since AdaIN focus on mean and variance, which are global information and lack of local feature, Liu et al.~\cite{liu2021adaattn} proposed AdaAttn, utilizing spatial attention score to dynamically calculate style feature statistics. For the purpose of efficiency, Zhu et al. introduced All-to-Key attention to reduce computational complexity and mitigate detail distortion.

\subsection{Motion Style Transfer}

Contrastive learning \cite{chen2020simple,ko2022self} plays an essential role in motion style transfer. Holden et al.~\cite{holden2016deep} adopts feedforward control network, achieving the first work in Motion style transfer. Since the effectiveness of AdaIN in image style transfer, Aberman et al. ~\cite{aberman2020unpaired} incorporate AdaIN in motion style transfer, along with deterministic autoencoders. Since Generative flow~\cite{dinh2014nice} has proven to be effective in the domain of image generation,  Wen et al.~\cite{wen2021autoregressive} introduced a generative flow based model, generating motions with less distortion. MoST, introduced by Kim et al.~\cite{kim2024most}, addressed style transfer between motions with differing contents by employing a transformer~\cite{vaswani2017attention} with part-attentive style modulation and Siamese encoders. Similarly, Guo et al.~\cite{guo2024generative} introduced a generative approach in the latent space, decomposing content and style motion into latent codes to enable flexible stylization. Motion Puzzle, proposed by Jang et al.~\cite{jang2022motion}, provides local style editing capabilities via revised AdaIN and attention mechanisms. Notably, the aforementioned works all employed AdaIN. The achievements of latent diffusion model~\cite{rombach2022high} have inspired its application in motion style transfer. Chen et al.~\cite{chen2023executing} proposed Motion Latent-based Diffusion (MLD), a diffusion process within a Variational AutoEncoder’s~\cite{kingma2013auto} latent space generates conditional human motions from inputs like text or action classes. MLD provides a foundational model for subsequent diffusion-based motion style transfer works. Zhong et al.~\cite{zhong2024smoodi} proposed SMooDi, which incorporated style guidance and a lightweight adaptor in MLD. Hu et al.~\cite{hu2024diffusion} fine-tune MLD with semantic-guided loss for few-shot style transfer using minimal style data. Song et al.~\cite{song2024arbitrary} extended MLD with multi-condition, including disentangling trajectory, content, and style. These works mentioned above are effective in generating stylized motions across varied content and styles.

\section{Method}

\subsection{Preliminary}

Given a style motion $M_S$ and a content motion $M_C$ as inputs, motion style transfer aims to disentangle style features from $M_S$ and apply them to $M_C$, thereby generating the stylized motion $M_G$. Existing motion stylization frameworks~\cite{kim2024most, aberman2020unpaired, guo2024generative, zhang2024generative} predominantly adopt Adaptive Instance Normalization (AdaIN), which is inherited image style transfer, as their core component. AdaIN achieves image style transfer by aligning the mean and variance of a content feature map with those of a style feature map, which can be formulated as:
\begin{equation}
    \mathrm{AdaIN}(x,y)=\sigma(y)\left(\frac{x-\mu(x)}{\sigma(x)}\right)+\mu(y).
\end{equation}

Though AdaIN aligns feature between style and content through mean and variance matching, this approach suffers inefficiency when applied to motion sequences. Unlike static images, motion sequence exhibits two intrinsic temporal properties which are asymmetry and dynamic. Asymmetry is reflected in irregular motion trajectories over time, while dynamics is manifested through varying joint speeds and accelerations. Using only the mean and variance is insufficient for capturing high-order and fine-grained style features of motion. Additionally, as presented in Figure \ref{fig:tsne},the result of t-SNE indicates significant differences in the distribution of statistics across various styles. Consequently, we incorporate skewness and kurtosis to enhance motion style transfer effectiveness, proposing out AStF, which will be detailed in following subsections.

\subsection{Method Overview}

\begin{figure*}[ht]
  \centering
  \includegraphics[width=\linewidth]{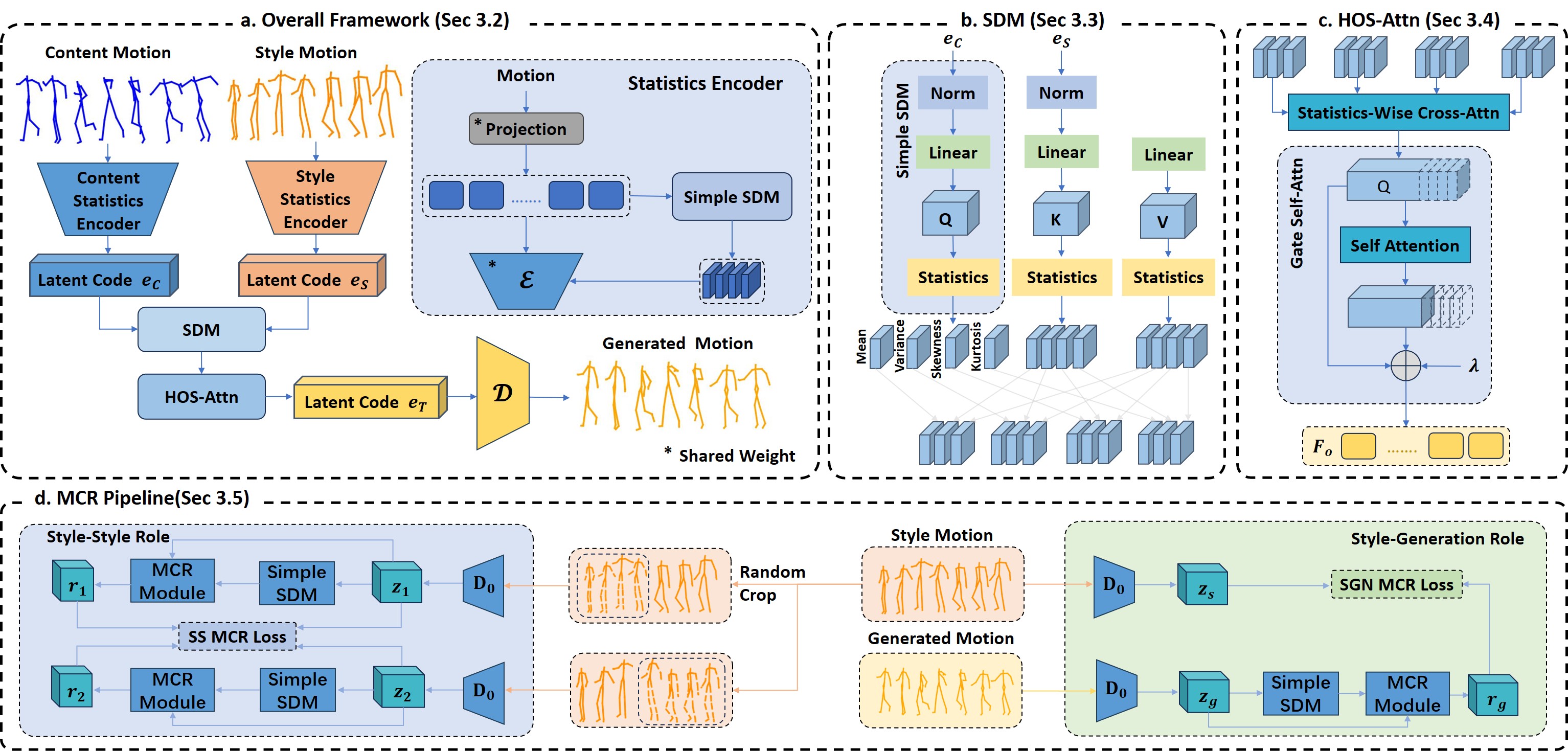}
  \caption{Method Overview. \textnormal{(a) Overall framework of AStF mainly comprising Content and Style Statistics-Encoder, Style Disentanglement Module (SDM), High-Order Multi-Statistics-Attention(HOS-Attn) and decoder. (b) Detailed structure of SDM and Simple SDM. (c) Detailed structure of HOS-Attn. (d) Pipeline of Motion Consistency Regularization (MCR) discriminator. }}
\label{fig:framework}
\end{figure*}

An overview of AStF pipeline is illustrated in Figure \ref{fig:framework}. Content motion `$M_C$ and style motion $M_S$ are first processed through Content Statistics Encoder and Style Statistics Encoder respectively, generating latent motion embeddings $e_S$ and $e_C$. In our Statistics Encoder, input motion is processed through projection layers to obtain motion embeddings. As previous works ~\cite{kim2024most, guo2024generative} utilize a learnable style token to aggregate the style features across an entire motion sequence, which could lead to a lack of style feature extraction capability, our Statistics Encoder replace it with the statistics features of motion embeddings extracted by Simple SDM, which will be detailed in section \ref{subsec:sdm}. We concatenate the statistics features with the corresponding embedding and pass the result to transformer ~\cite{vaswani2017attention} encoder, generating latent code $e_S$ and $e_C$. Subsequently, these two embeddings are input into SDM, projecting $e_S$ and $e_C$ to query $Q$, key $K$ and value $V$, then calculate four statistics consist of mean, variance, skewness and kurtosis, represented by $\mu$, $\sigma^2$, ${\gamma}$ and ${\beta}$ respectively. This process is formulated as:
\begin{equation}
\mathcal{SDM}:\left(e_S,e_C;\theta_\mathcal{SDM}\right)\mapsto(\mu, \sigma^2, {\gamma},{\beta}),
  \label{eq:sdm}
\end{equation}
where $\theta_\mathcal{SDM}$ denote the learnable parameter of SDM. Subsequently, four types of statistics are input into HOS-Attn along with $Q$ to perform Statistics-wise Cross Attention and Gate Self Attention, which ultimately results in transferred motion embedding $e_T$. The HOS-Attn process is formulated as:
\begin{equation}
  \mathcal{HOS}:\left(\mu, \sigma^2, {\gamma},{\beta};\theta_\mathcal{HOS}\right)\mapsto(e_T),
  \label{eq:hos}
\end{equation}
where $\theta_\mathcal{HOS}$ denote the learnable parameter of HOS-Attn. After acquiring $e_T$, the motion decoder is utilized to decode the motion embedding $e_T$ into the desired motion sequence $M_G$.

\subsection{Style Disentanglement Module}
\label{subsec:sdm}
Unlike conventional approaches that rely solely on mean and variance to align motion style distributions, we propose a Style Disentanglement Module by introducing skewness and kurtosis as higher-order statistics, which we named as SDM. In Statistical Analysis, skewness and kurtosis are higher-order statistics that describe the shape of a probability distribution. Skewness measures the asymmetry of a probability distribution, indicating whether data are concentrated on one side. Kurtosis quantifies the tailedness and peakedness of a distribution, revealing whether extreme values are more frequent or data are more dispersed. Specifically, as presented in Figure \ref{fig:framework} (b), we defined Simple SDM and SDM. Simple SDM is implemented in the Statistic Encoder to replaced the learnable style token utilized in previous works ~\cite{kim2024most, guo2024generative}, achieving adaptive style features extraction. SDM take $e_c$ and $e_s$ as input, following instance normalization and linear to obtain the key triplet $(Q, K, V)$, representing query, key and value. We define mean as $\mu$, variance as $\sigma^2$, skewness as $\gamma$, kurtosis as $\beta$, we utilize the following definitions and formulas to calculate each statistics of $(Q, K, V)$:

\begin{equation}
  \left\{
    \begin{aligned}
      \mu &= \frac{1}{F} \sum_{f=1}^{F} x_{d,f,j} , \\
      \sigma^2 &= \frac{1}{F} \sum_{f=1}^{F} (x_{d,f,j} - \mu)^2 , \\
      \gamma &= \frac{1}{F} \sum_{f=1}^{F} \left( \frac{x_{d,f,j} - \mu}{\sigma} \right)^3 , \\
      \beta &= \frac{1}{F} \sum_{f=1}^{F} \left( \frac{x_{d,f,j} - \mu}{\sigma} \right)^4 ,
    \end{aligned}
  \right.
  \label{eq:all_aligned}
\end{equation}
where $d$, $f$, and $j$ indicate feature dimensions, frames, and joints, respectively. This calculation for each $(Q, K, V)$ tensors yield 12 independent statistics (e.g., $\mu_Q$,$\sigma^2_Q$, $\gamma_Q$, $\beta_Q$ for $Q$). These statistics are then classified into four domain-specific statistics groups: $[\mu_Q,\mu_K,\mu_V]$, $[\sigma^2_Q,\sigma^2_K,\sigma^2_V]$, $[\gamma_Q,\gamma_K,\gamma_V]$, $[\beta_Q,\beta_K,\beta_V]$. This structured aggregation enables explicit modeling of both hierarchical dynamics and cross-tensor correlations, providing a compact and expressive representation for motion style transfer.

\subsection{High-Order Multi-Statistics-Attention}

To holistically fuse the statistical characteristics of motion features and enhance the correlation between content and style embeddings, we propose HOS-Attn. As presented in Figure \ref{fig:framework} (c), HOS-Attn explicitly models four domain-specific statistics groups extracted from the SDM through inter-domain interaction and cross-domain interaction. For each domain-specific statistics group, HOS-Attn computes its cross-domain interaction through Statistics-wise Cross-Attention, which applies dedicated cross-attention to each group of statistics, outputting refined statistics features $\mu_c$,$\sigma^2_c$, $\gamma_c$, $\beta_c$. The refined statistics features are subsequently concatenated with $Q$ to form an augmented feature. This augmented feature is then processed by cross-domain interaction through Gate Self-Attention to integrate global statistical features. Inside the Gate Self-Attention, we first apply self-attention to the augmented feature, then discard the concatenated statistical terms, resulting in output $F_c$. To preserve the integrity of the original motion semantics while incorporating statistical guidance, a gating residual mechanism is introduced to produce the final output $F_o$:

\begin{equation}
    F_{o}=\lambda_g\times F_c \oplus (1-\lambda_g)\times Q,
  \label{gate-residual}
\end{equation}
where $\oplus$ indicated to element-wise addition. The similarity coefficient $\lambda_g$ is calculated based on cosine similarity. The formula of $\lambda_g$ is as follows:

\begin{equation}
    \lambda_g=\frac{\frac{Q\cdot K}{\|Q\|\|K\|}+1}{2} .
\end{equation}

\subsection{Motion Consistency Regularization}
In motion style transfer, prior works equip the discriminator with only a coarse style-classification head, leading to the issue of style fading which is characterized by weak expression and temporal degradation of style in the generated motion. Our analysis suggests the root cause lies in the discriminator’s reliance on global features while ignoring local features, allowing the generator to "cheat". Inspired by Ko et al. ~\cite{ko2022self}, we introduce Motion Consistency Regularization (MCR) to enhances the discriminator with two complementary roles, style-style and style-generation, to address this problem. As presented in Figure \ref{fig:framework}(d), in style-style role, MCR splits a style motion into two subsequences to evaluate internal coherence. In style-generation role, MCR computes the similarity between style and generated motion to capture global features. This design effectively mitigates the problem of style fading. 

Specifically, MCR discriminator consists of a feature extractor $\text{D}_0$ followed by a classification head $\text{D}_1$. By taking style motion $M_S$ as input, style-style role first apply random crop to $M_S$, separating $M_S$ into two subsequences $s_1$ and $s_2$. Then we obtain the intermediate feature $z_1$ by $z_1=\text{D}_0{(s_1)}$, and similarly obtain $z_2$. Subsequently, we further process $z_1$ and $z_2$ by Simple SDM and MCR Module, a lightweight module comprising two linear layers with LeakyReLU activation, to obtain refined features $r_1$ and $r_2$. We compute the MCR loss between $r_1$ and $z_2$ within their valid temporal region $\Omega$ on the basis of cosine similarity. This process is formulated as:

\begin{equation}
\mathrm{sim}(r_1,z_2;\Omega)\equiv\sum_{(f)\in\Omega}-\frac{r_1[f]}{\|r_1[f]\|_2}\cdot\frac{z_2[f]}{\|z_2[f]\|_2},
\end{equation}
where $\|\cdot\|_{2}$indicates $\ell_{2}$-norm, $f$ indicates to frame index. The style-style MCR loss is formulated as:

\begin{equation}
\label{eq:ss}
    \mathcal{L}_{\mathrm{ss}}=\frac{1}{3}\mathrm{sim}_{\mathrm{nc}}(r_1,\text{F}_\text{sg}(z_2))+\frac{1}{3}\mathrm{sim}_{\mathrm{nc}}(r_2,\text{F}_\text{sg}(z_1)),
\end{equation}
where $\text{F}_\text{sg}(\cdot)$ indicates to stop-gradient. For the style generation component, the generated motion \( M_G \) is passed through \( \text{D}_0 \) to obtain the corresponding feature representation \( z_g \). Similarly, the style reference motion \( M_S \) is processed by the same module to extract the feature \( z_s \). The feature \( z_g \) is subsequently passed through the Simple SDM and MCR modules to obtain the refined representation \( r_g \). This facilitates the computation of the style generation MCR loss, defined as follows:

\begin{equation}
\label{eq:sgn}
    \mathcal{L}_{\mathrm{sgn}}=\frac{1}{3}\mathrm{sim}_{\mathrm{nc}}(r_g,\text{F}_\text{sg}(z_s)).
\end{equation}

\subsection{Loss Function}
\label{subsec:loss}
The training objective of our AStF is defined through loss function for the discriminator $\mathcal{L}_\text{D}$ and the generator $\mathcal{L}_\text{G}$. As adversarial loss and R1 loss \cite{mescheder2018training} are proven to be effective in the training of discriminator in previous works ~\cite{kim2024most, jang2022motion, aberman2020unpaired} , we combined our MCR loss, which consists of style-style MCR loss and generation-style MCR loss in equation \ref{eq:ss} and \ref{eq:sgn}, with adversarial loss and R1 loss to generate our discriminator loss:
\begin{equation}  
\label{eq:disc}
\mathcal{L}_\text{D}=\mathcal{L}_\mathrm{adv}^\text{D}+\mathcal{L}_\mathrm{R1}+\lambda_\text{MCR}\cdot(\mathcal{L}_\mathrm{ss}+\mathcal{L}_\mathrm{sgn}),
\end{equation}
where $\lambda_\text{MCR}$ is the hyperparameter, setting $\lambda_\text{MCR}=1$ to balance the contributions. $\mathcal{L}_\text{G}$ integrates multiple constraints for content-style disentanglement, which are commonly used in existing methods, including adversarial ($\mathcal{L}_\text{adv}^\text{G}$), reconstruction ($\mathcal{L}_\text{r}$), content cycle consistency ($\mathcal{L}_\text{cc}$) and style cycle consistency ($\mathcal{L}_\text{cs}$) losses. However, during training, we observed that $M_G$ tends to adhere to $M_C$, indicating insufficient style feature extraction and transfer. To address this issue, we propose a style-align loss $\mathcal{L}_\text{a}$ that processes $M_G$ through an encoder to obtain its latent representation $e_g$, then computes an L2 loss between $e_g$ and the style latent code $e_s$. This mechanism effectively guides the style motion to properly embed its stylistic characteristics into the $M_G$. Consequently, our $\mathcal{L}_\text{G}$ is formulated as follows:

\begin{equation}
\label{eq:gen}
    \mathcal{L}_\text{G}=\mathcal{L}_\text{adv}^\text{G}+\lambda_\text{r}\mathcal{L}_\text{r}+\lambda_\text{c}(\mathcal{L}_\text{cc}+\mathcal{L}_\text{cs})+ \lambda_\text{a}\mathcal{L}_\text{a},
\end{equation}
where $\lambda_{*}$ denotes the weight of each loss. We set $\lambda_\text{r}=3$, $\lambda_\text{c}=3$, $\lambda_\text{a}=1$ in our experiment.

\begin{table*}[t]
\centering
\caption{Quantitative results on the test set of BFA ~\cite{aberman2020unpaired} and Xia ~\cite{xia2015realtime} datasets. \textnormal{Style FID $\downarrow$, Cnt FID $\downarrow$, Sty Acc $\uparrow$, Cnt Acc $\uparrow$, and Geo Dis $\downarrow$ denote style/content FID, style/content accuracy, and geodesic distance respectively. Bold font indicates the best result.}}
\label{tab:quan}
\begin{tabular*}{\textwidth}{@{\extracolsep{\fill}}lccc|ccccc@{}}
\toprule
\multirow{2}{*}{Method} & \multicolumn{3}{c|}{BFA dataset} & \multicolumn{5}{c}{Xia dataset} \\
\cmidrule(lr){2-4} \cmidrule(lr){4-9}
 & \begin{tabular}{@{}c@{}}Style \\ FID $\downarrow$\end{tabular} & \begin{tabular}{@{}c@{}}Style \\ Acc $\uparrow$\end{tabular} & \begin{tabular}{@{}c@{}}Geo \\ Dis $\downarrow$\end{tabular} & \begin{tabular}{@{}c@{}}Style \\ FID $\downarrow$\end{tabular} & \begin{tabular}{@{}c@{}}Content \\ FID $\downarrow$\end{tabular} & \begin{tabular}{@{}c@{}}Style \\ Acc $\uparrow$\end{tabular} & \begin{tabular}{@{}c@{}}Content \\ Acc $\uparrow$\end{tabular} & \begin{tabular}{@{}c@{}}Geo \\ Dis $\downarrow$\end{tabular} \\
\midrule
Aberman et al. \cite{aberman2020unpaired}  & 5.412 & 0.525 & 0.631 & 0.602 & 0.00574 & 0.377 & 0.575 & 0.869 \\
MotionPuzzle \cite{jang2022motion} & 3.487 & 0.745 & 0.431 & 0.285 & 0.00541 & 0.764 & 0.573 & 0.568 \\
Park et al. \cite{park2021diverse} & 3.501 & 0.731 & 0.422 & 0.272 & 0.00574 & 0.775 & 0.481 & 0.767 \\
MoST \cite{kim2024most}  & 3.107 & 0.775 & 0.350 & 0.188 & 0.00151 & 0.699 & 0.867 & \textbf{0.398} \\
GenMoStyle \cite{guo2024generative}  & 1.892 & 0.845 & 0.434 & 0.192 & 0.00147 & 0.873 & 0.817 & 0.435 \\
\textbf{AStF (ours)} & \textbf{1.558} & \textbf{0.905} & \textbf{0.414} & \textbf{0.157} & \textbf{0.00101} & \textbf{0.930} & \textbf{0.903} & 0.440 \\
\bottomrule
\end{tabular*}
\end{table*}

\section{Experiments}
In this section, we evaluate our proposed AStF and analyze its essential characteristics.

\subsection{Experimental Settings}

\textbf{Experimental Implementation.} For a fair comparison, we follow the experimental protocol of GenMoStyle \cite{guo2024generative} and take different style motions and content motions as input to analyze the performance of our method. We train our models on an NVIDIA A6000 48 GB. We optimize the generator and discriminator using separate Adam optimizers ~\cite{kingma2014adam} with the same $\beta_1=0.9$ and $\beta_2=0.99$, and different learning rates of $1\times10^{-5}$ and $1\times10^{-6}$, respectively. We set the batch size as 16 and trained for about 400K iterations. To verify the effectiveness of the proposed model, we compare it with other state-of-art methods in motion style transfer: Aberman et al. \cite{aberman2020unpaired}, MotionPuzzle proposed by Jang et al. \cite{jang2022motion}, Park et al. \cite{park2021diverse}, and MoST proposed by Kim et al. \cite{kim2024most}, GenMoStyle proposed by Guo et al. ~\cite{guo2024generative}.

\textbf{Datasets.} Following previous works, we evaluated our AStF on two publicly available motion datasets, Xia dataset ~\cite{xia2015realtime}and BFA dataset ~\cite{aberman2020unpaired}. Xia dataset ~\cite{xia2015realtime} contains 8 distinct motion styles, including angry, childlike, and strutting, etc, and we categorize the contents into 6 classes, including walking, jumping, kicking, etc.. BFA dataset ~\cite{aberman2020unpaired} consists of 16 motion styles, however, not be labeled by content. Both datasets have 31-joint skeleton, and we select 21 key joints for training and evaluation, following the approach proposed by Aberman et al. ~\cite{aberman2020unpaired}. Given that each motion sequence in the Xia dataset varies in frame length, we downsample each sequence by a factor of two, then apply zero-padded until each motion sequence reaches a fixed frame length $L_m$, which we set to 200. For the BFA dataset, we directly split each Biovision Hierarchy (BVH) file into groups of 400 frames, discarding any remaining samples with fewer than 400 frames. We then apply downsampling by a factor of two, resulting in each sample having a fixed frame length of 200. 

\textbf{Evaluation Metrics.} For the purpose of effectively demonstrating the robustness of our proposed methodology, we utilize five separate evaluation metrics to quantify both content retention and style fidelity of the generated motion with the input of content motion and style motion. We employ the content and style Fr\'{e}chet Inception Distance (FID)~\cite{heusel2017gans}, a widely accepted metric in previous work, to quantify the feature distribution similarity between the input motions and generated motion. We utilize recognition accuracy to measure semantic consistency. Additionally, we compute the geodesic distance \cite{guo2024generative} between the 3D rotation matrices of content motion and generated motion. For the calculation of content and style FID, we first trained a content classifier and a style classifier on the train set of Xia dataset and BFA dataset. The content classifier and the style classifier are applied to compute content and style accuracy.

\begin{table*}
  \caption{Ablation experiment results} 
  \label{tab:ablation}
  \begin{tabular}{lccccc}
    \toprule
    Model Variants & Style FID $\downarrow$ & Content FID $\downarrow$ & Style Acc $\uparrow$ & Content Acc $\uparrow$ & Geo Dis $\downarrow$ \\
    \midrule
    w/o Entire MCR Loss & 0.204 & 0.00130 & 0.723 & 0.856 & 0.452 \\
    w/o Style-Style MCR Loss & 0.188 & 0.00118 & 0.782 & 0.881 &  0.437 \\
    w/o Style-Generation MCR Loss & 0.191 & 0.00111 & 0.773 & 0.889 & 0.423 \\
    \midrule
    w/o Simple SDM & 0.203 & 0.00222 & 0.710 & 0.851 & 0.343 \\
    \midrule
    w/o Style Align Loss & 0.353 & 0.00218 & 0.637 & 0.838 & \textbf{0.245} \\
    \midrule
    w/o Skew & 0.166 & 0.00113 & 0.805 & 0.847 & 0.430 \\
    w/o Kurt & 0.160 & 0.00135 & 0.799 & 0.832 & 0.447 \\
    w/o Skew and Kurt & 0.179 & 0.00168 & 0.718 & 0.857 & 0.450 \\
    \midrule
    $\lambda_\text{MCR}=0.5$ & 0.198 & 0.00151 & 0.837 & 0.876 & 0.471 \\
    $\lambda_\text{MCR}=2$ & 0.186 & 0.00194 & 0.875 & 0.801 & 0.523 \\
    $\lambda_\text{MCR}=3$ & 0.207 & 0.00217 & 0.804 & 0.735 & 0.551 \\
    $\lambda_\text{a}=0.5$ & 0.189 & 0.00128 & 0.871 & 0.901 & 0.416 \\
    $\lambda_\text{a}=2$ & 0.175 & 0.00187 & 0.892 & 0.812 & 0.462 \\
    $\lambda_\text{a}=3$ & 0.161 & 0.00198 & 0.919 & 0.784 & 0.501 \\
    \midrule
    \textbf{AStF (Full Model)} & \textbf{0.157} & \textbf{0.00115} & \textbf{0.930} & \textbf{0.903} & 0.440 \\    
    \bottomrule
  \end{tabular}
\end{table*}

\begin{table*}
  \caption{Qualitative Results on Xia ~\cite{xia2015realtime} and BFA ~\cite{aberman2020unpaired} datasets. \textnormal{Please refer to the red indications: dashed circles and arrows represent undesired content, while solid lines indicates to the style that needs to be transferred. }}
  \label{tab:qualitative}
  \begin{tabular}{lcccc}
    \toprule
      & (1)  & (2)  & (3) & (4) on BFA dataset \cite{aberman2020unpaired} \\
    \midrule
    Content motion & 
        \parbox[c]{3cm}{\centering Childlike jump \\ \includegraphics[width=3cm]{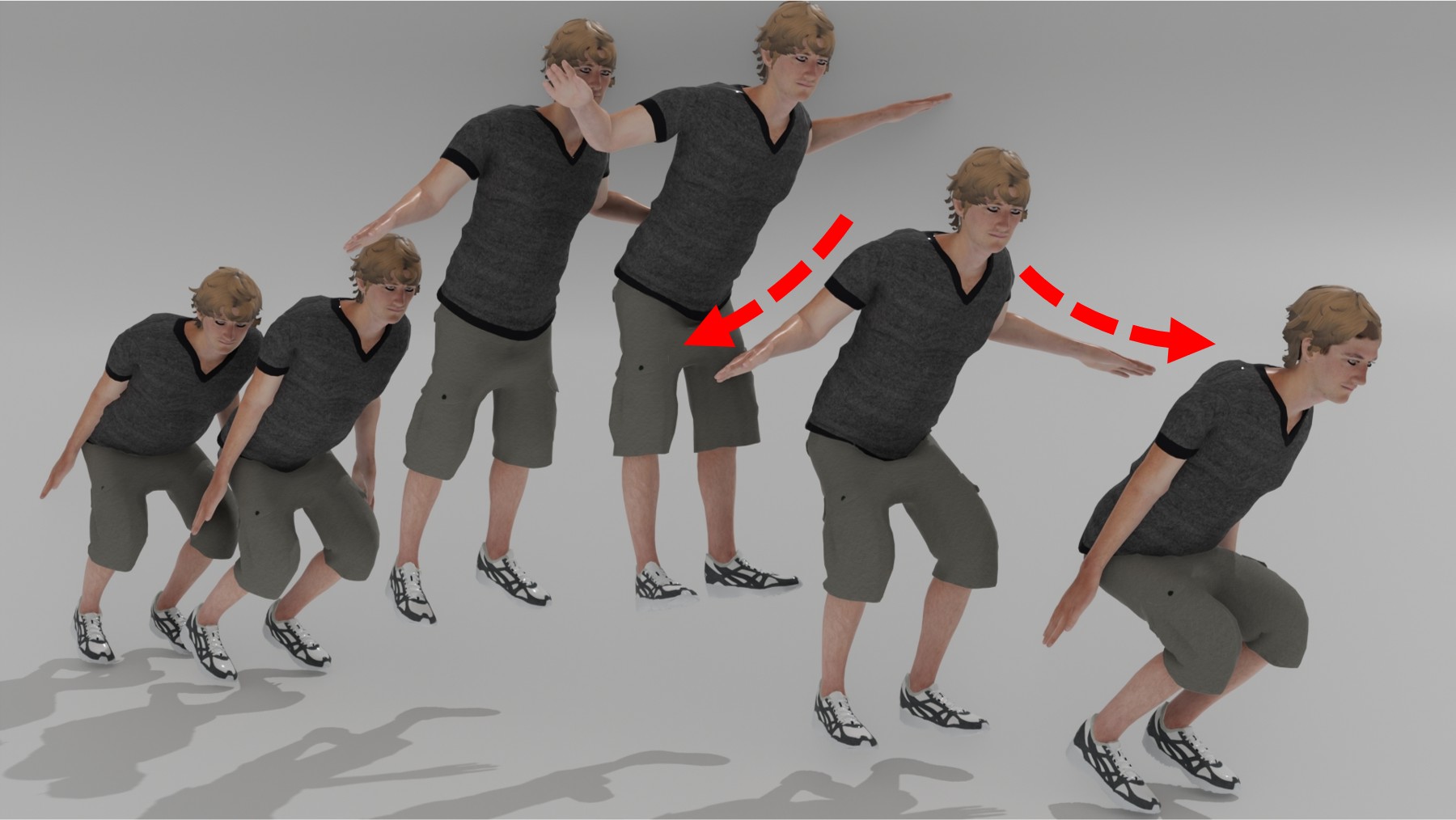}} & 
        \parbox[c]{3cm}{\centering Strutting walk \\ \includegraphics[width=3cm]{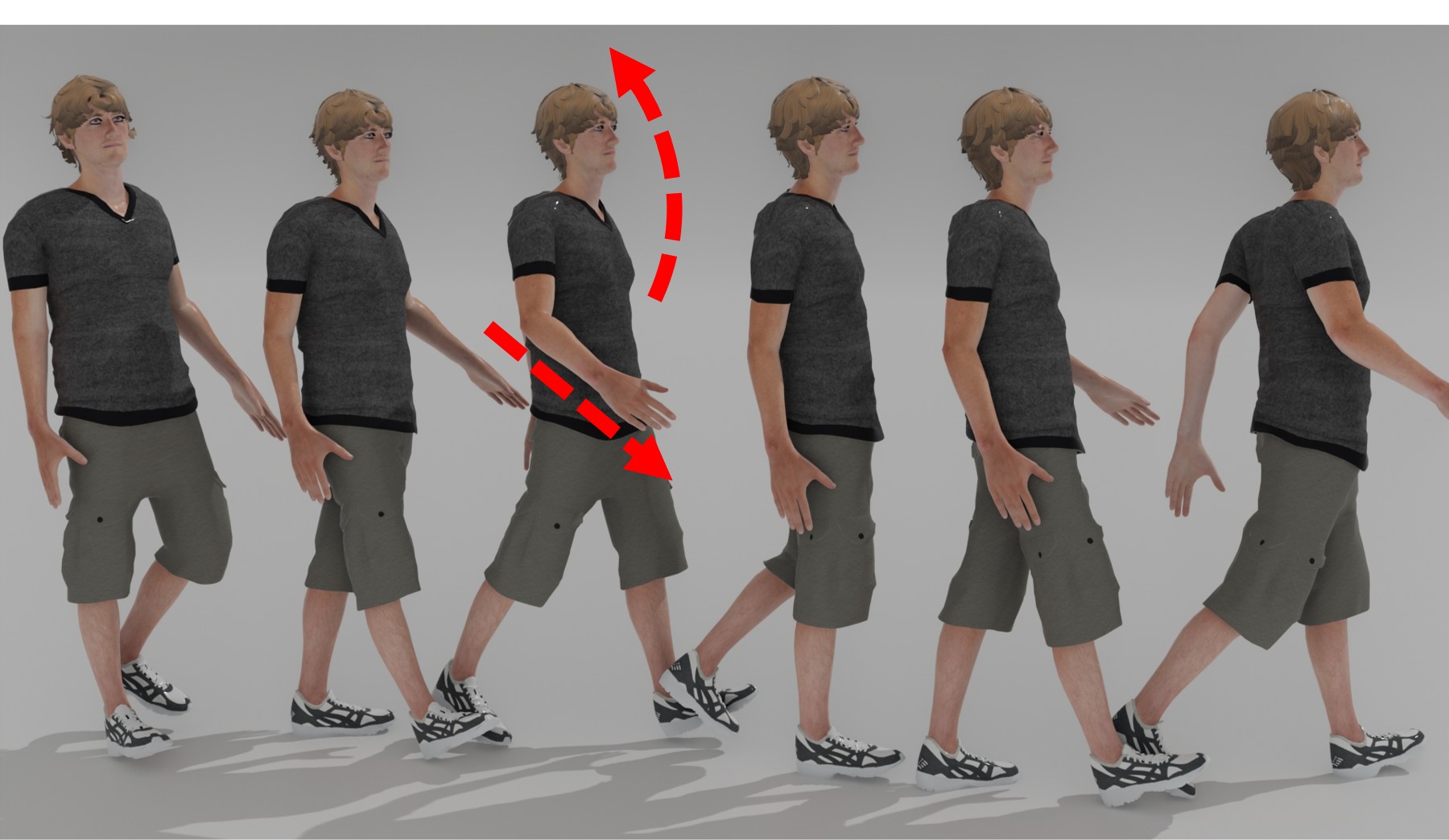}} & 
        \parbox[c]{3cm}{\centering Sexy kick \\  \includegraphics[width=3cm]{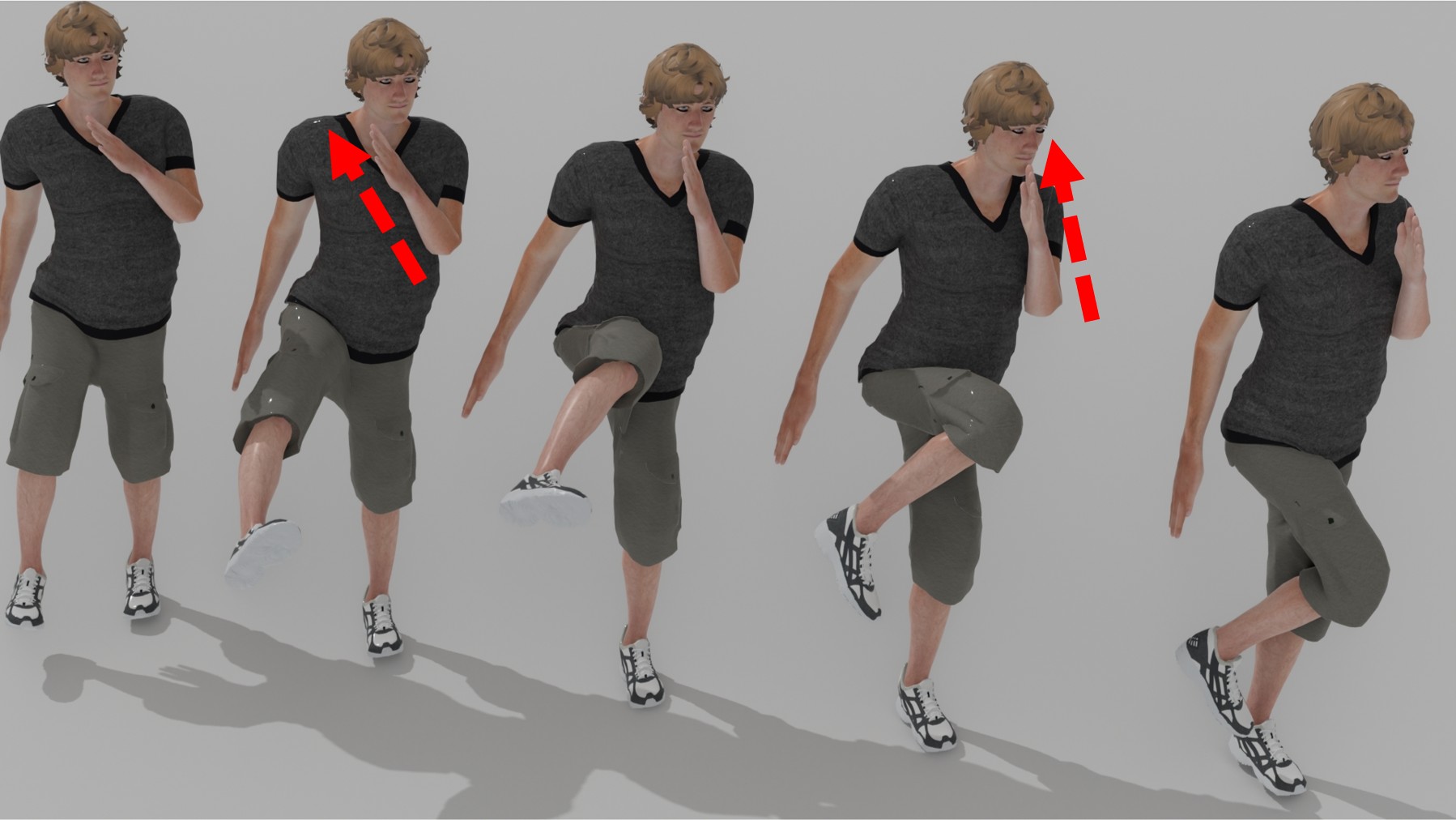}}  &
        \parbox[c]{3cm}{\centering Hurried \\ \includegraphics[width=3cm]{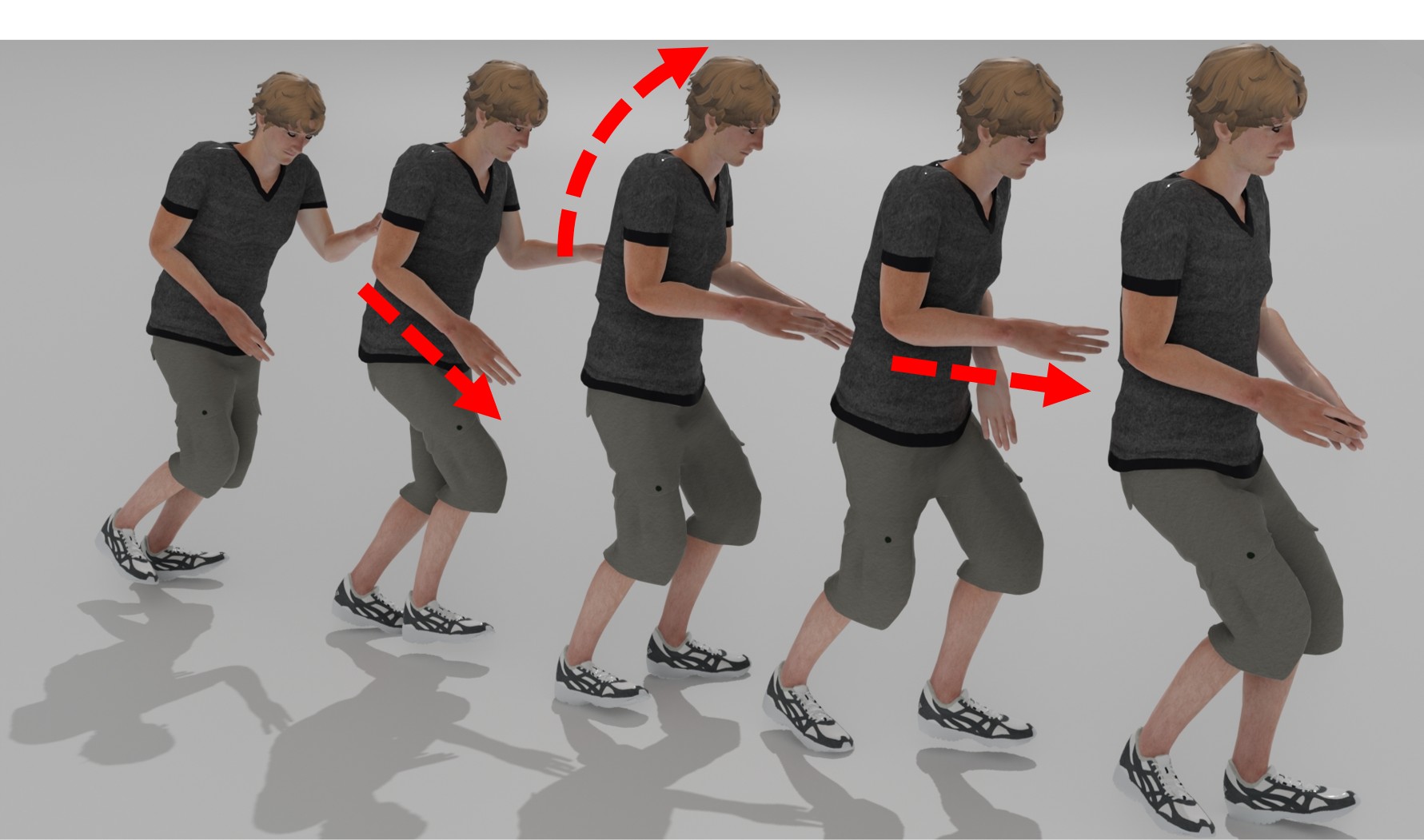}}   \\ 
    Style motion & 
        \parbox[c]{3cm}{\centering Depressed walk \\ \includegraphics[width=3cm]{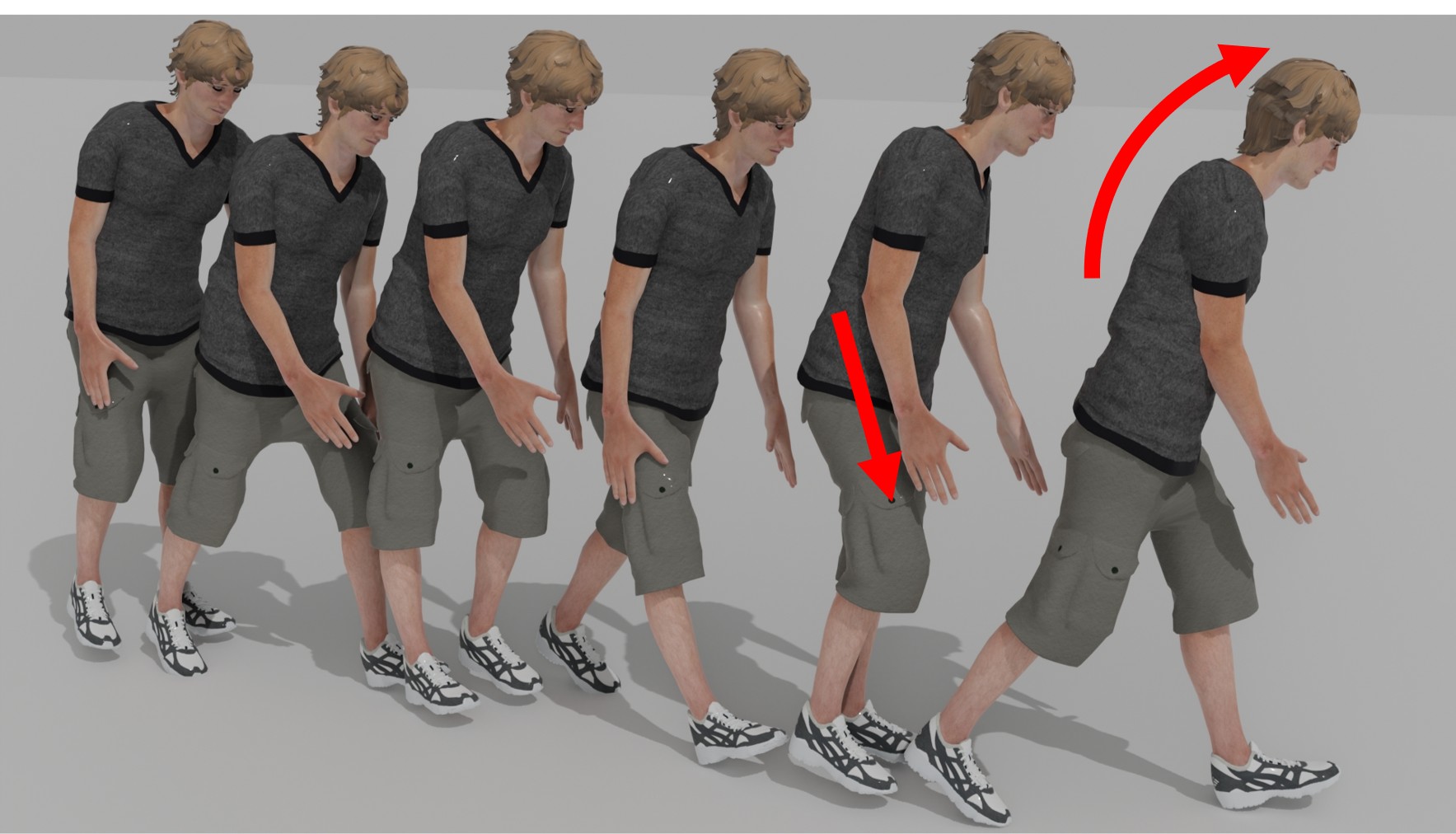} }  &
        \parbox[c]{3cm}{\centering Childlike walk \\  \includegraphics[width=3cm]{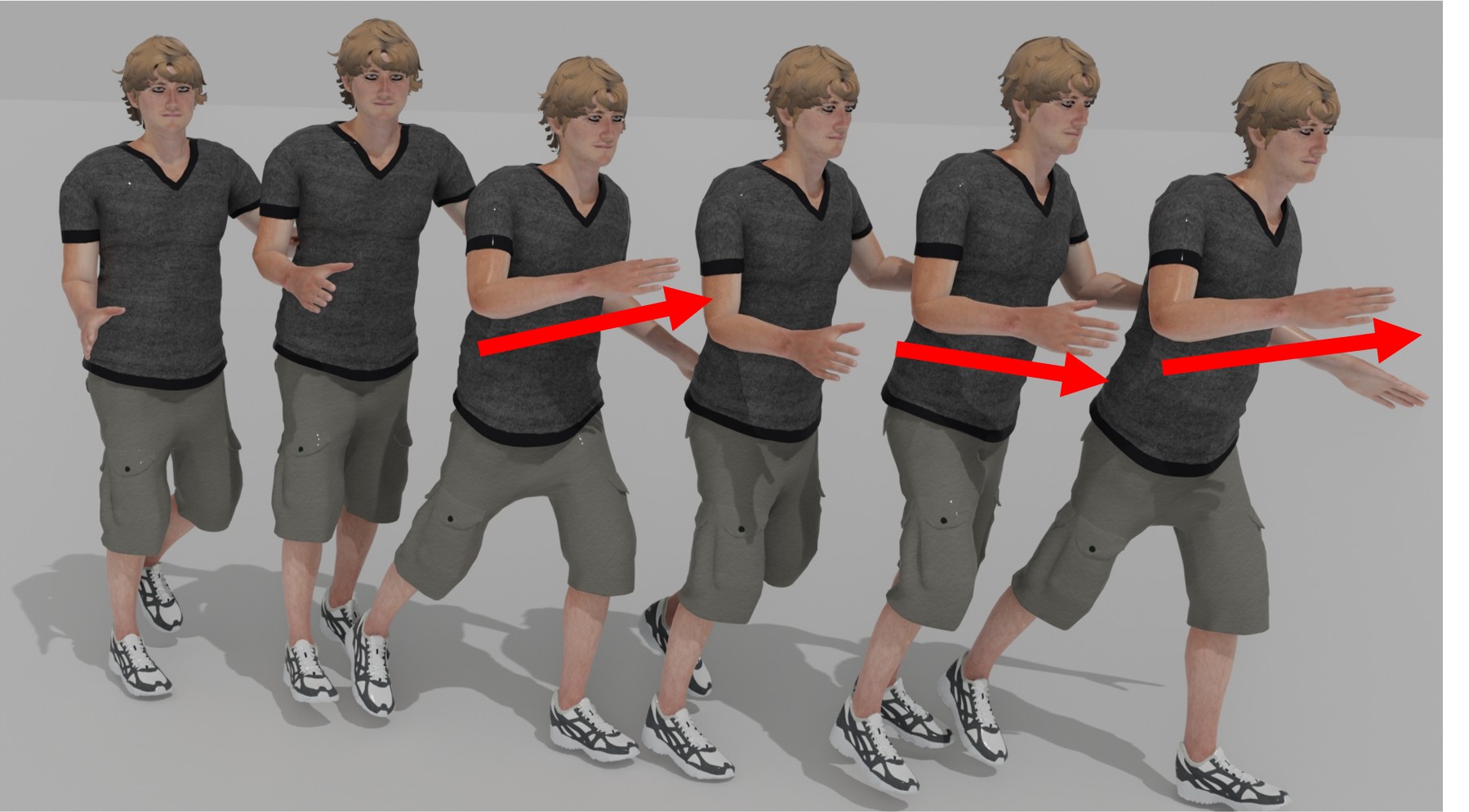}}  &
        \parbox[c]{3cm}{\centering Angry jump \\  \includegraphics[width=3cm]{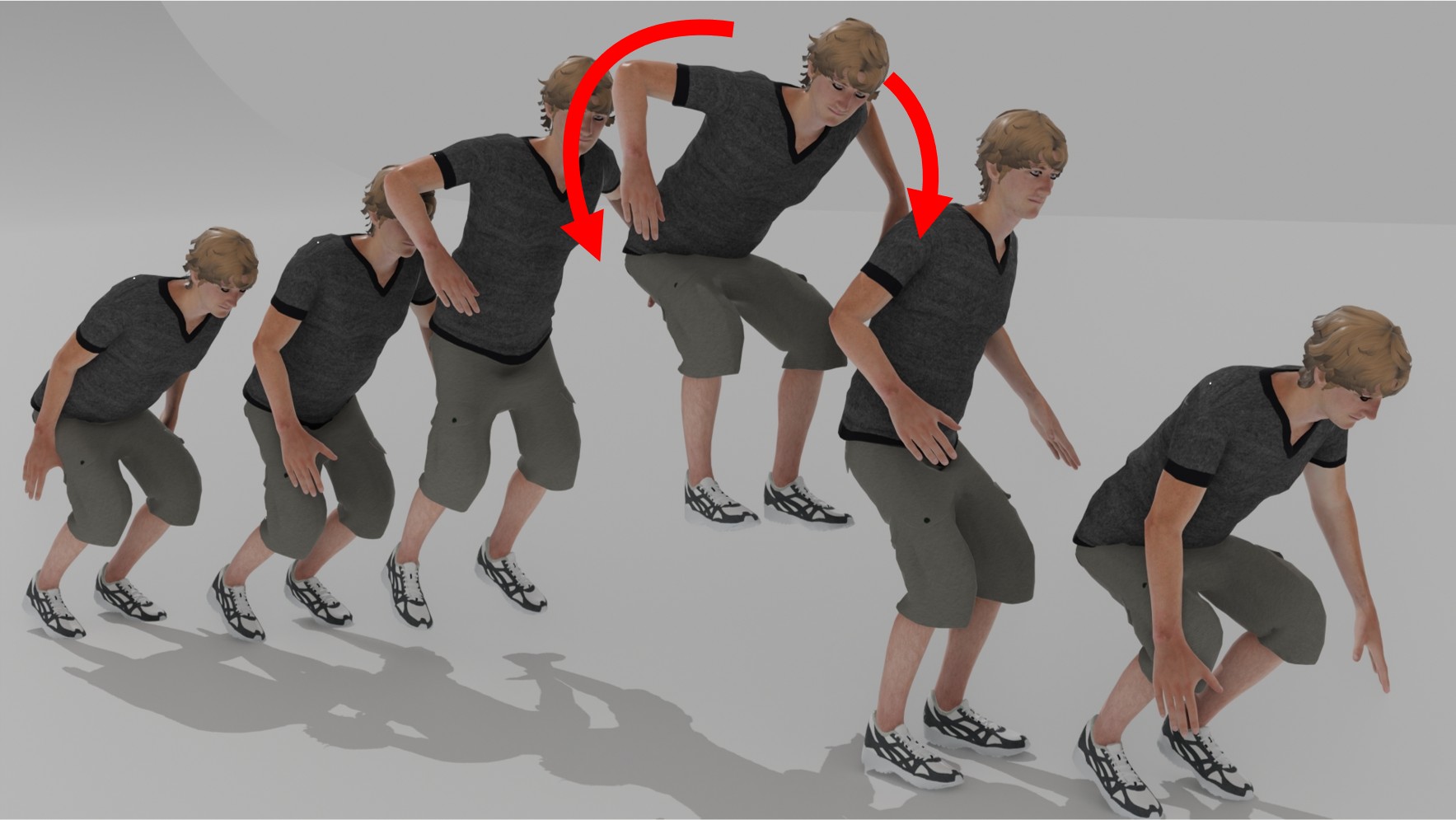} }  &
        \parbox[c]{3cm}{\centering Proud \\  \includegraphics[width=3cm]{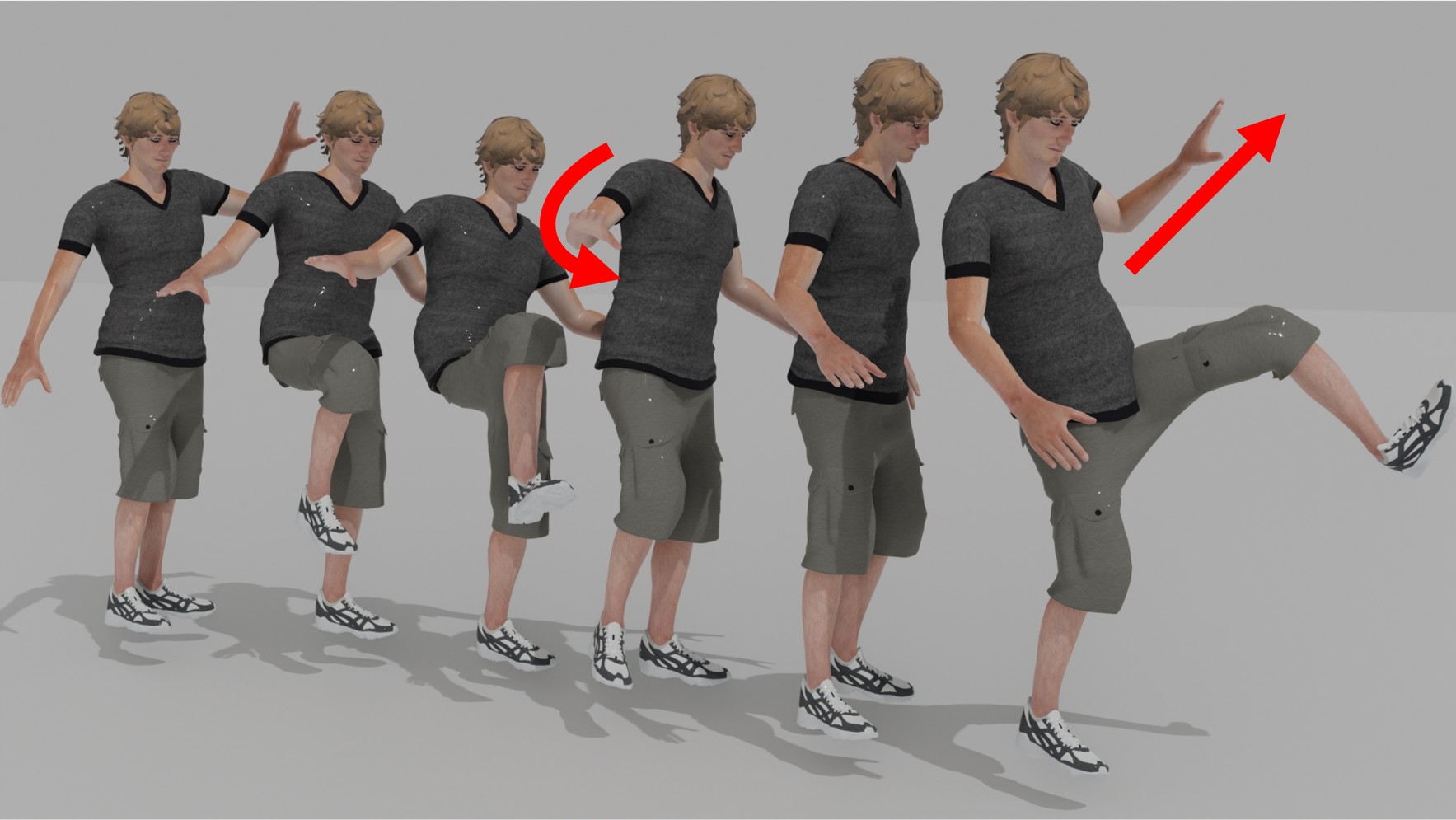}}    \\
    \midrule
    \textbf{AStF(Ours)} & 
        \parbox[c]{3cm}{\centering  \includegraphics[width=3cm]{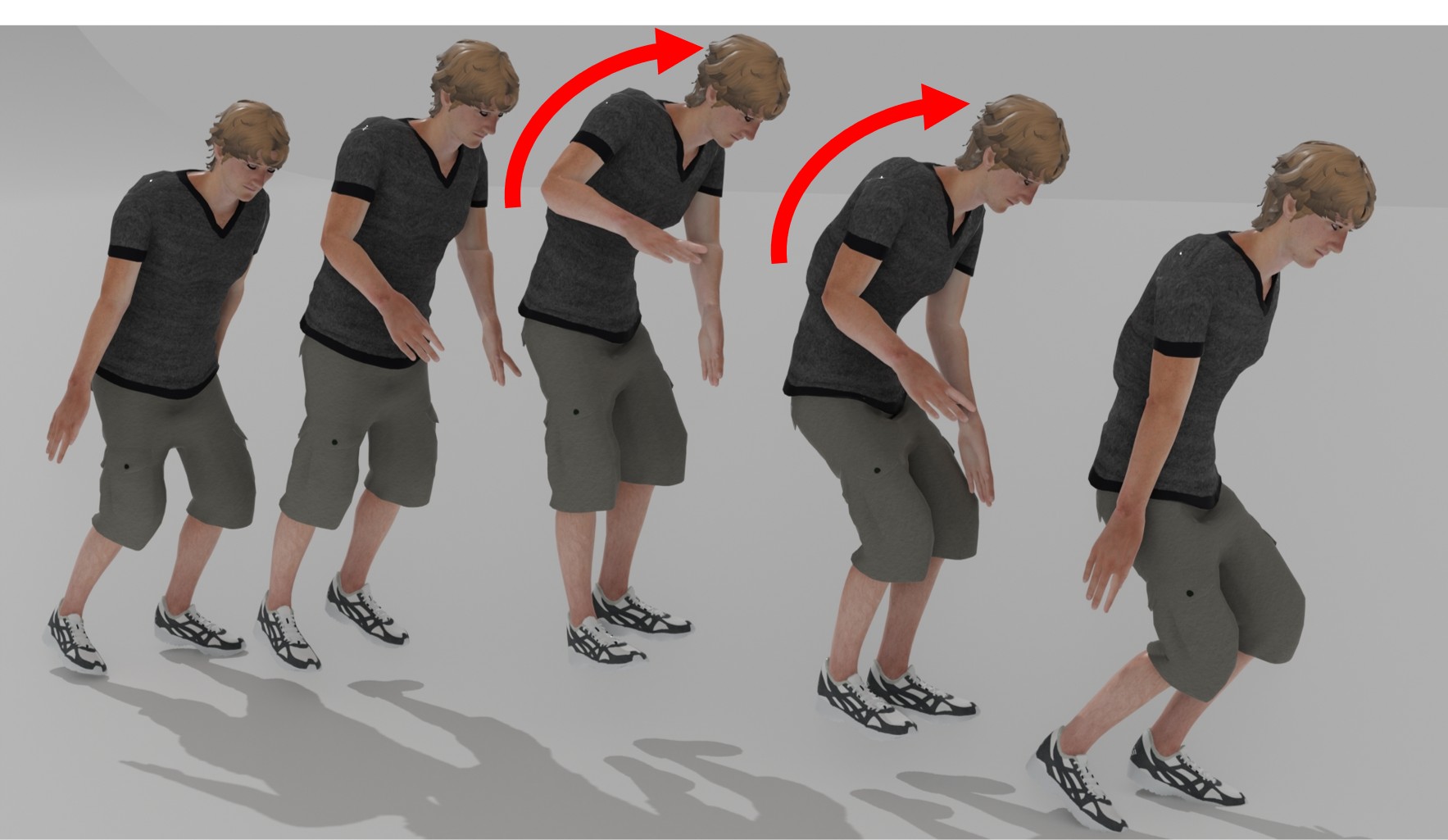} }  &
        \parbox[c]{3cm}{\centering  \includegraphics[width=3cm]{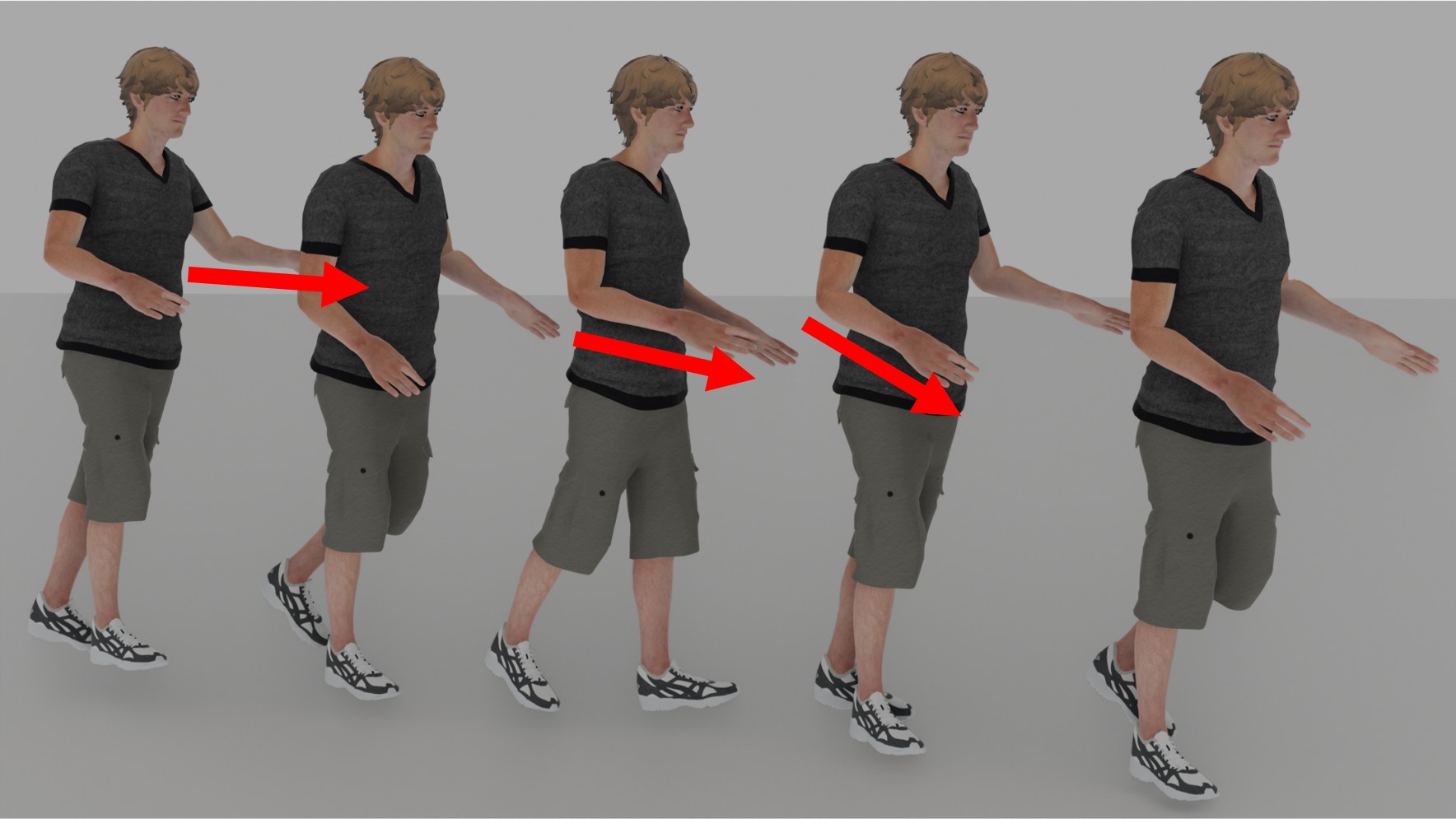}}  &
        \parbox[c]{3cm}{\centering  \includegraphics[width=3cm]{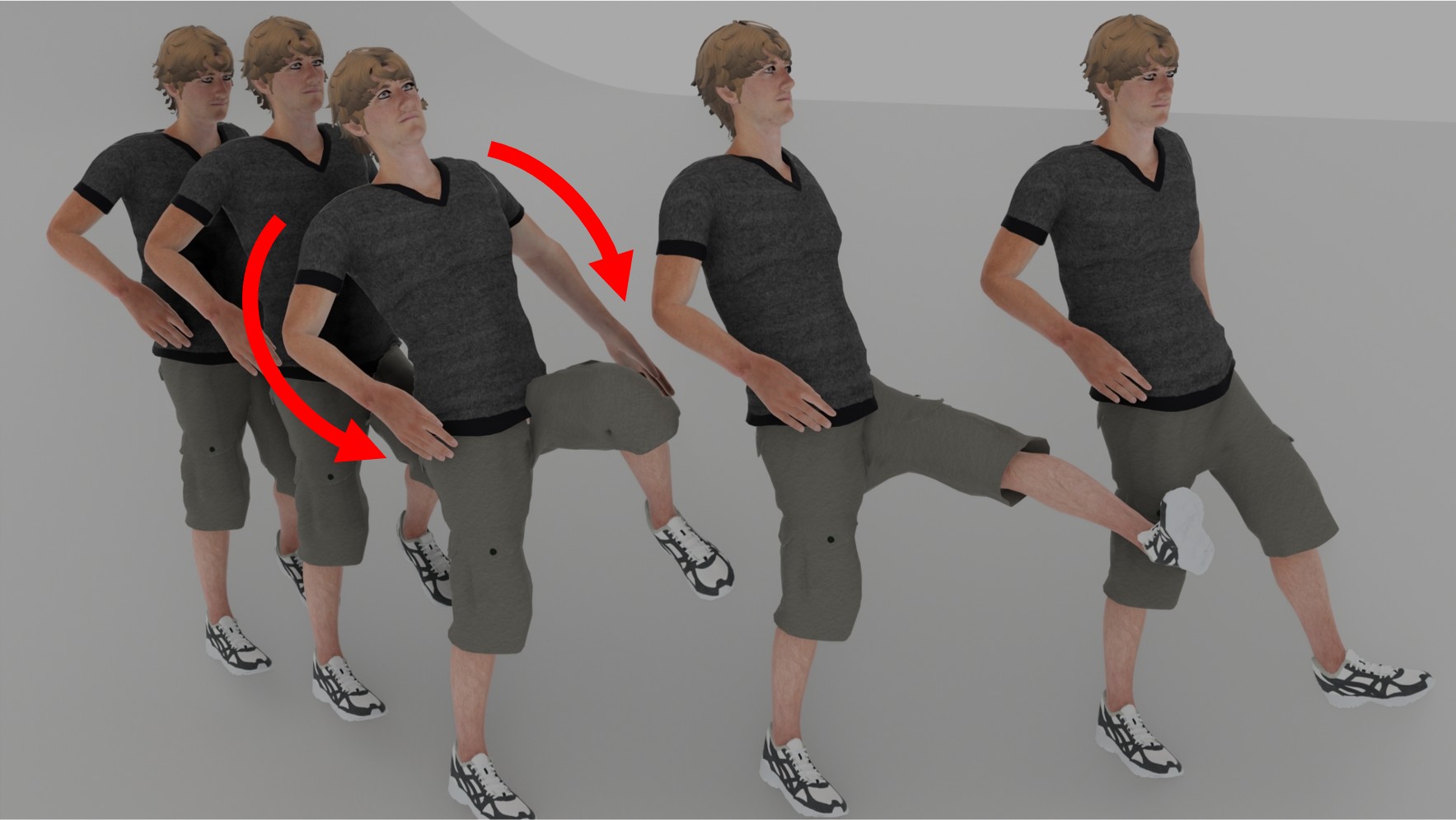} } &
        \parbox[c]{3cm}{\centering  \includegraphics[width=3cm]{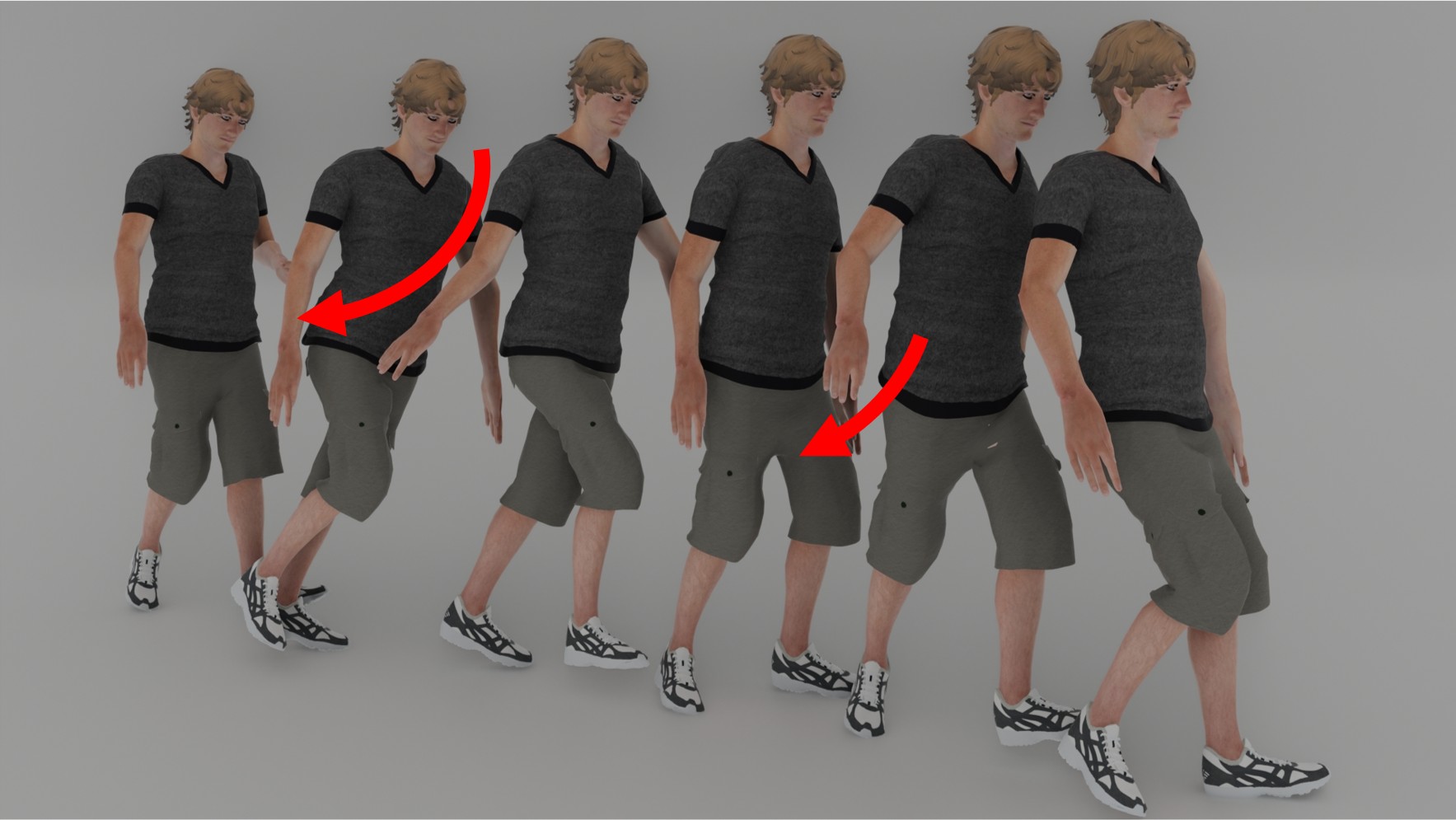}}  \\ 
    \midrule
    MoST \cite{kim2024most} & 
        \parbox[c]{3cm}{\centering \includegraphics[width=3cm]{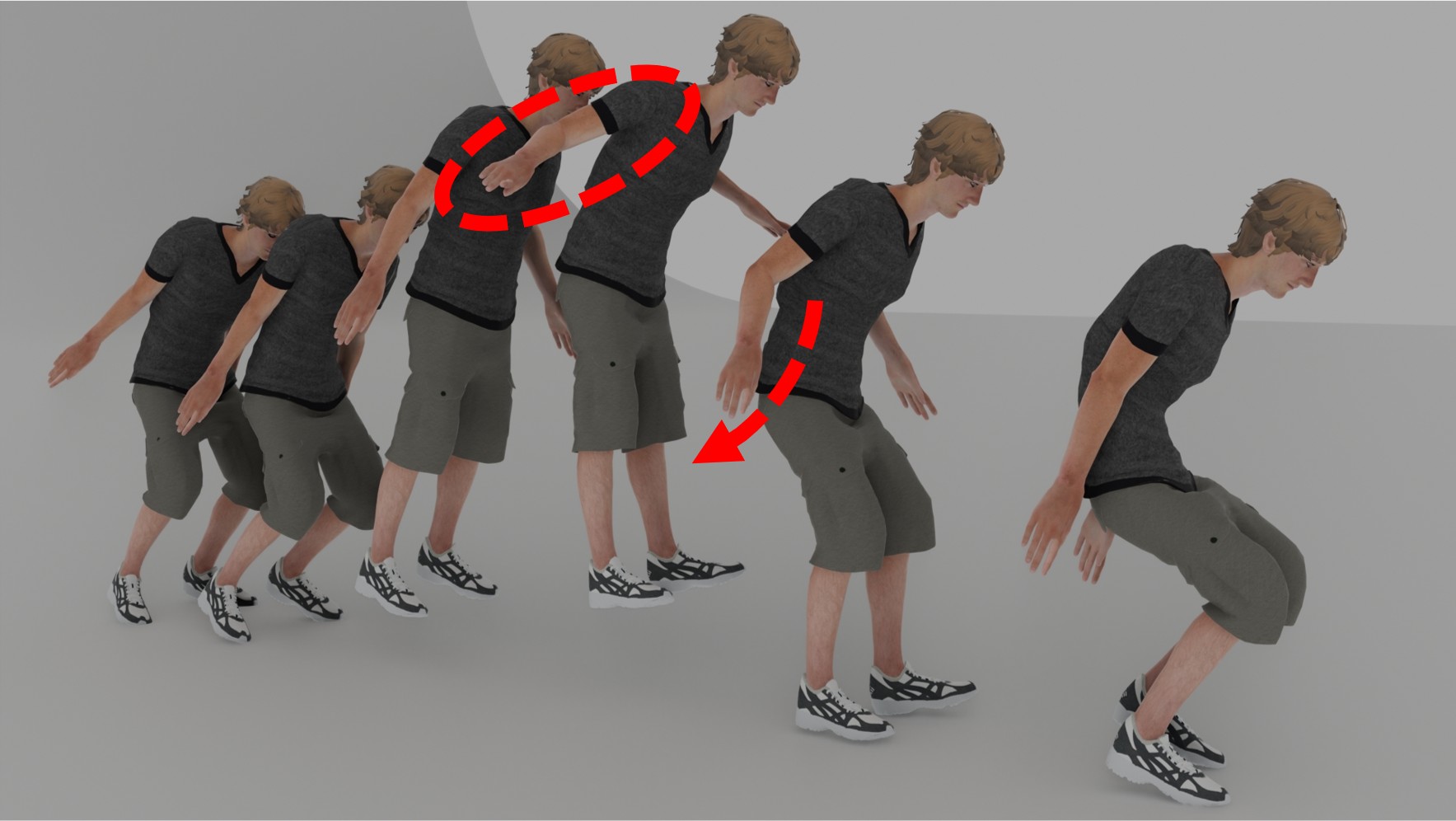} }  &
        \parbox[c]{3cm}{\centering \includegraphics[width=3cm]{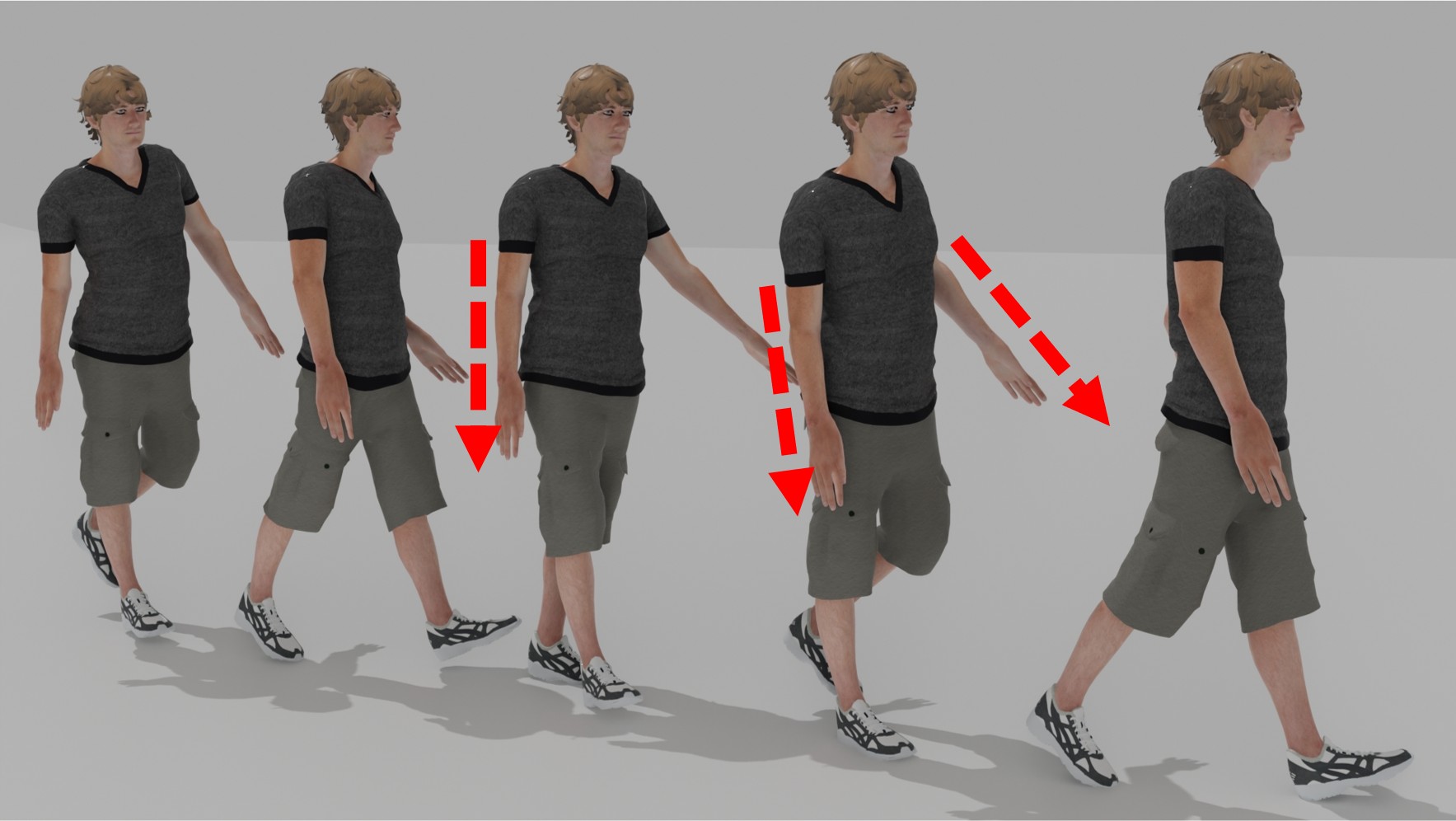}}  &
        \parbox[c]{3cm}{\centering \includegraphics[width=3cm]{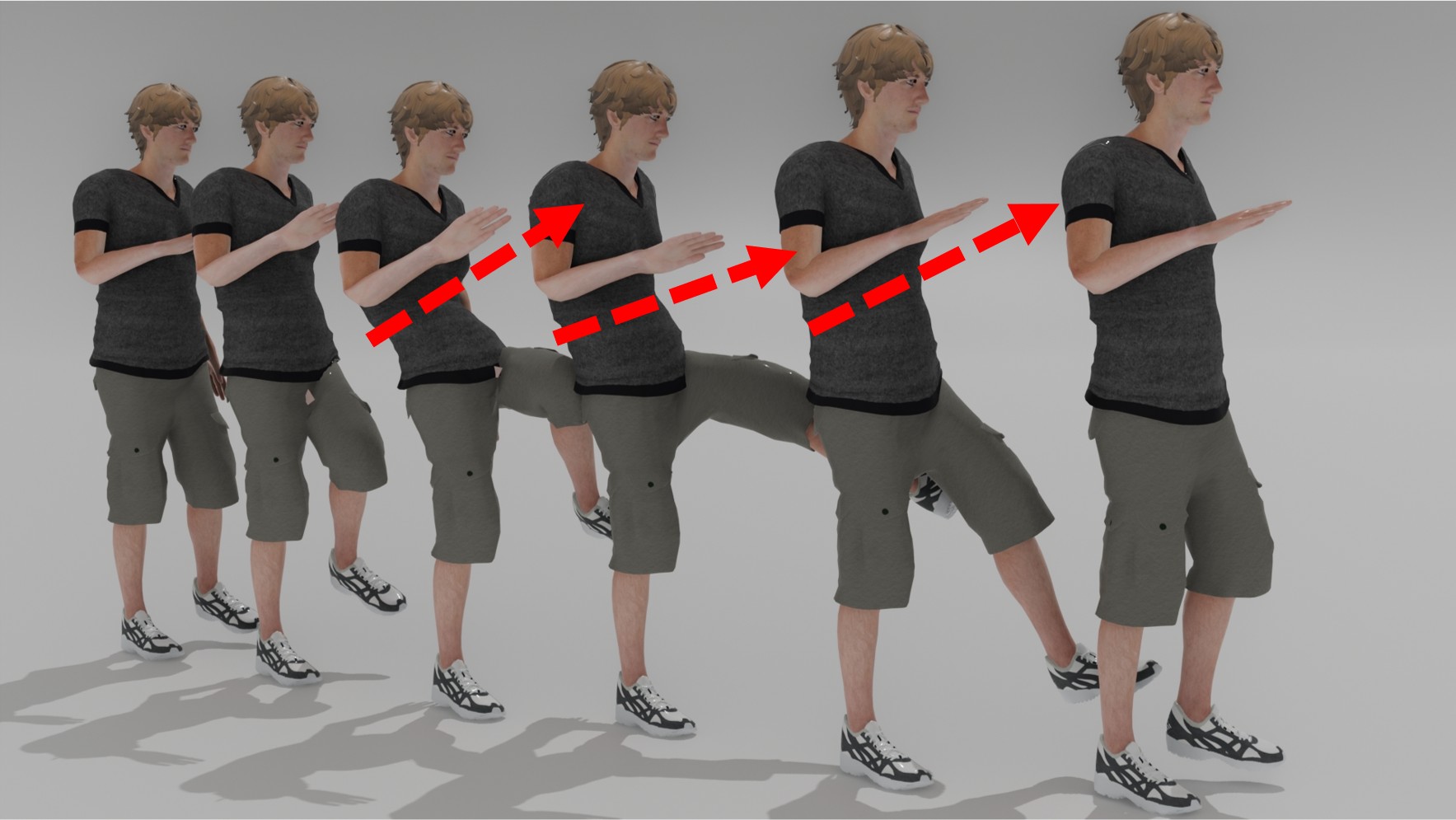} } &
        \parbox[c]{3cm}{\centering \includegraphics[width=3cm]{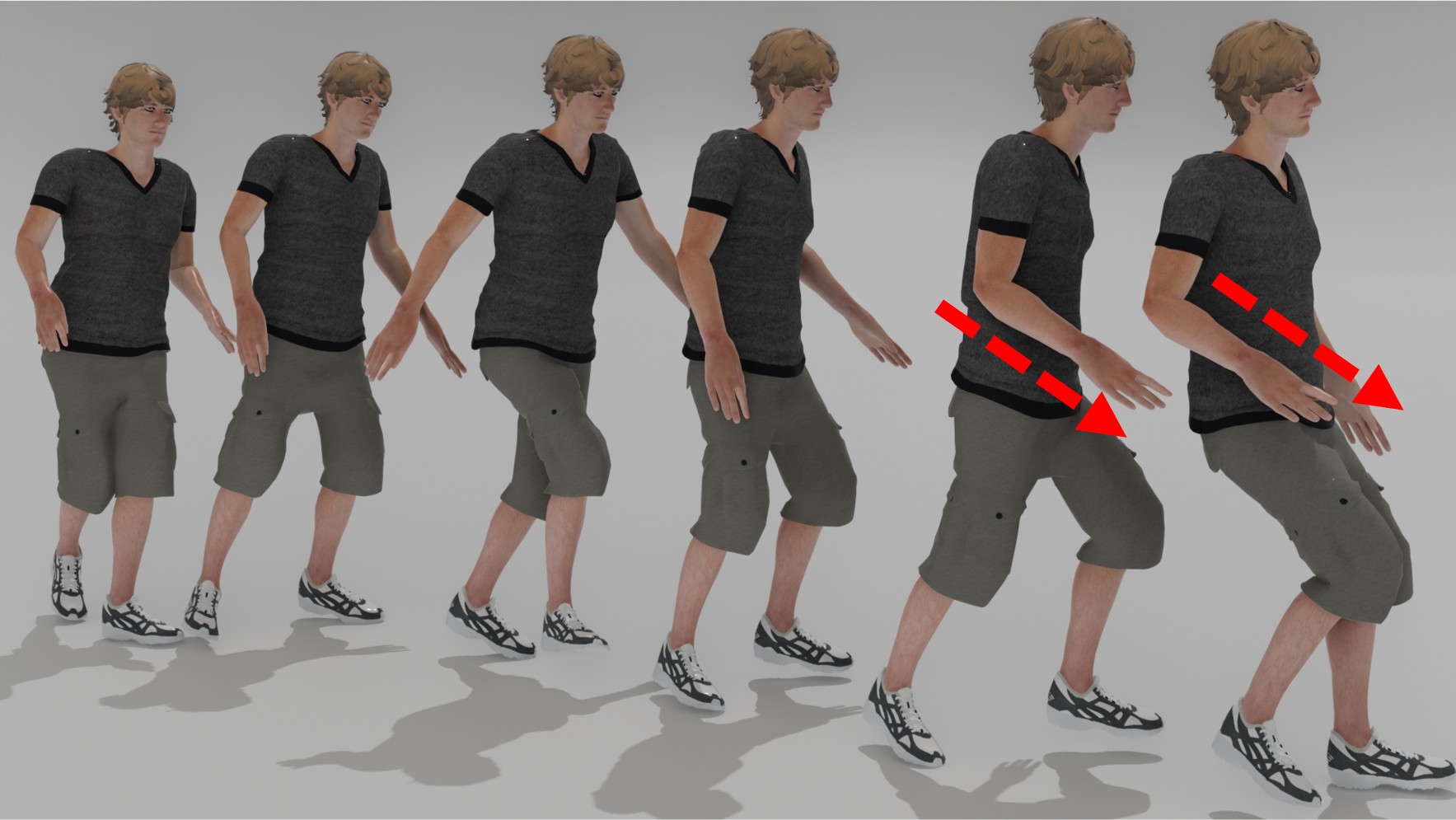}}  \\ 
    \midrule
    GenMoStyle ~\cite{guo2024generative} &
        \parbox[c]{3cm}{\centering  \includegraphics[width=3cm]{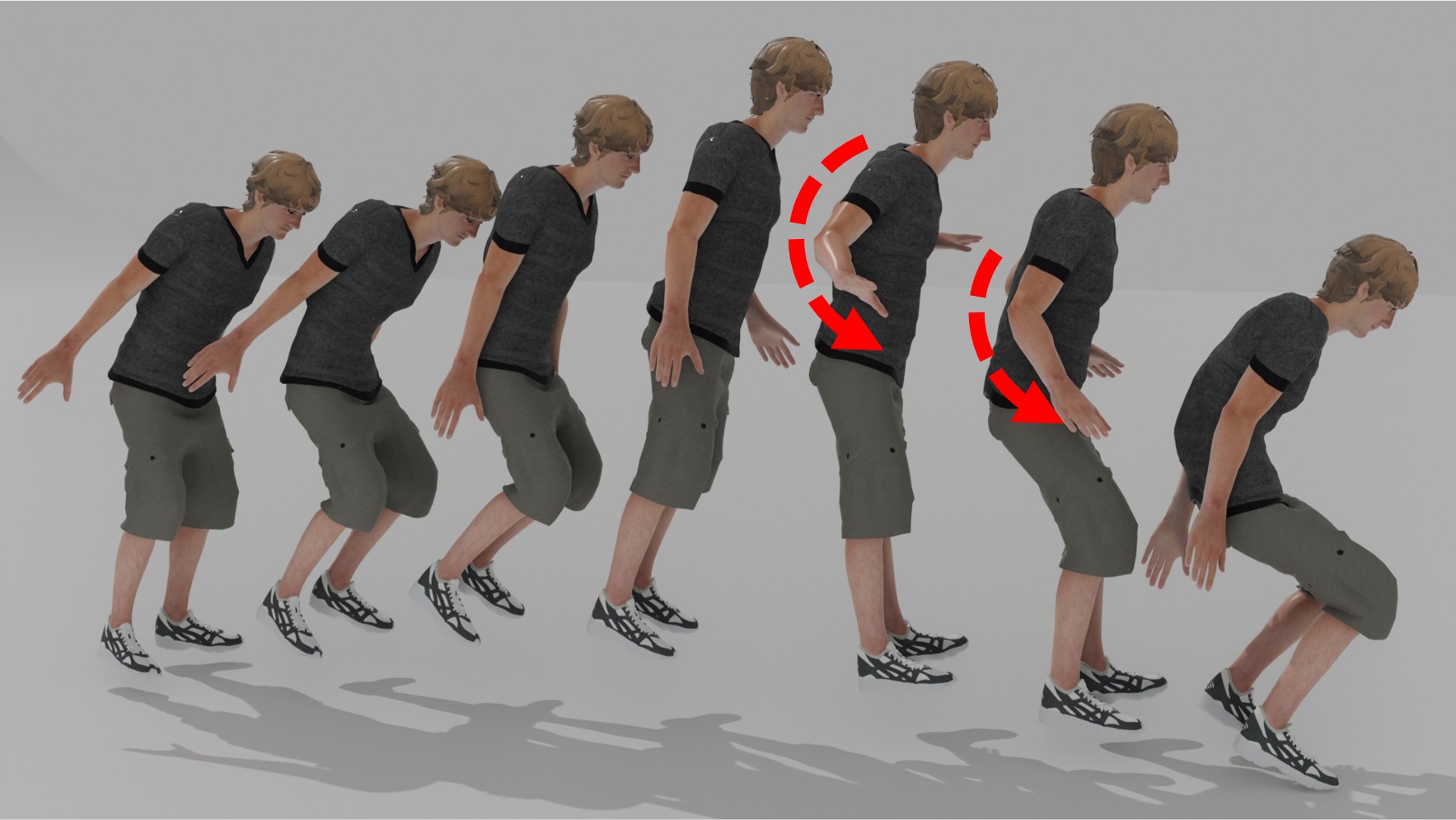} }  &
        \parbox[c]{3cm}{\centering  \includegraphics[width=3cm]{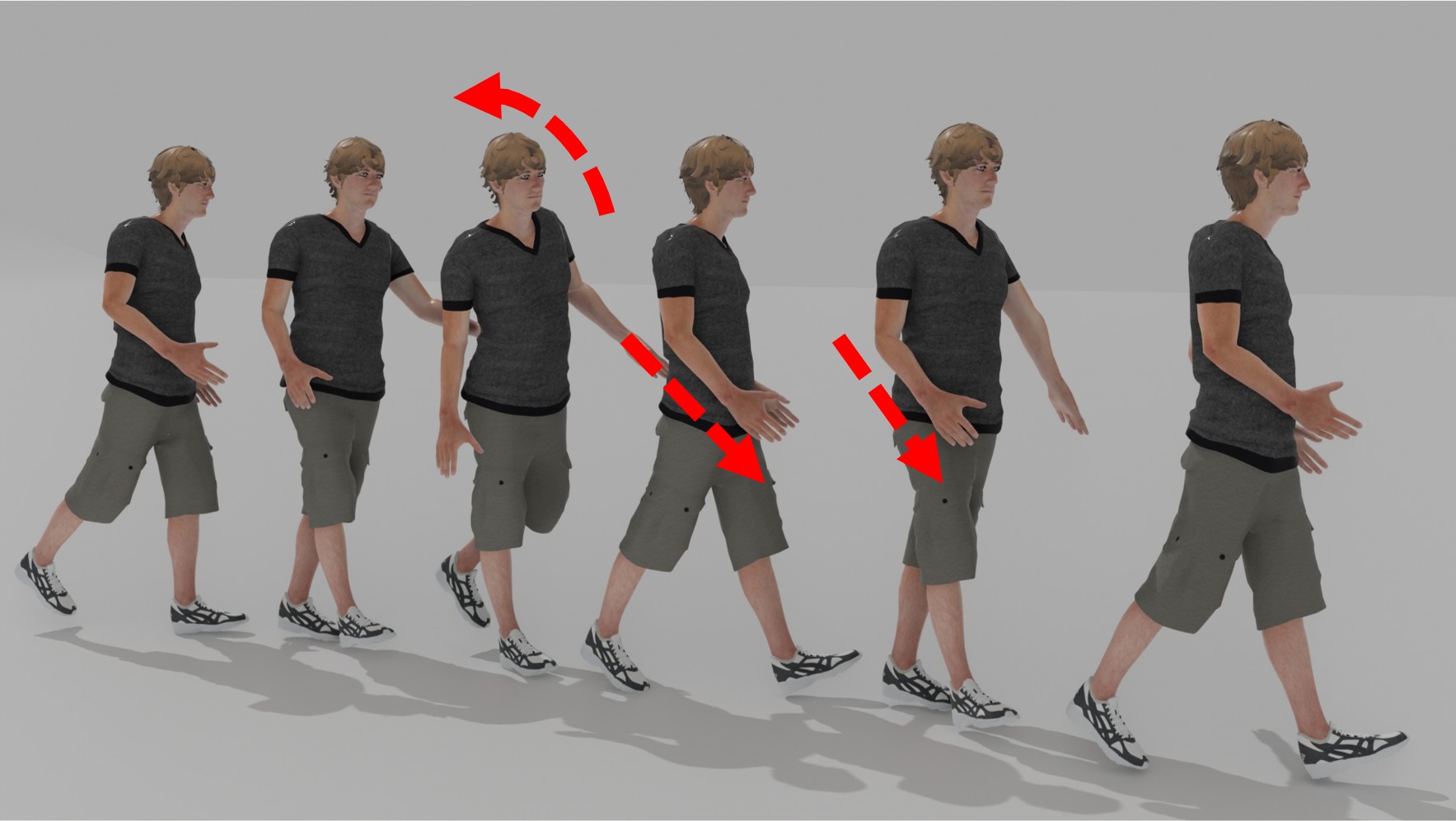}}  &
        \parbox[c]{3cm}{\centering  \includegraphics[width=3cm]{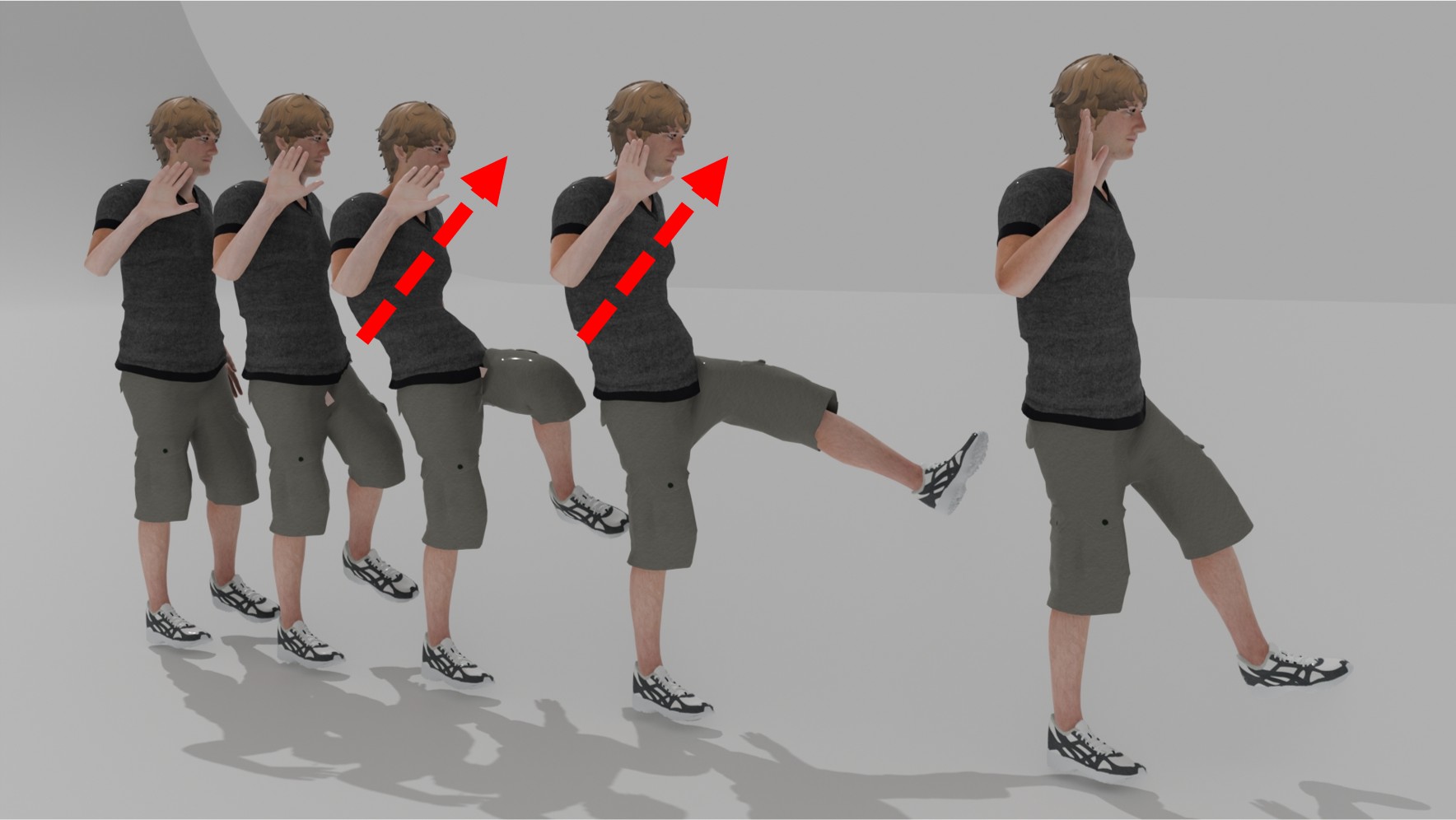} } &
        \parbox[c]{3cm}{\centering  \includegraphics[width=3cm]{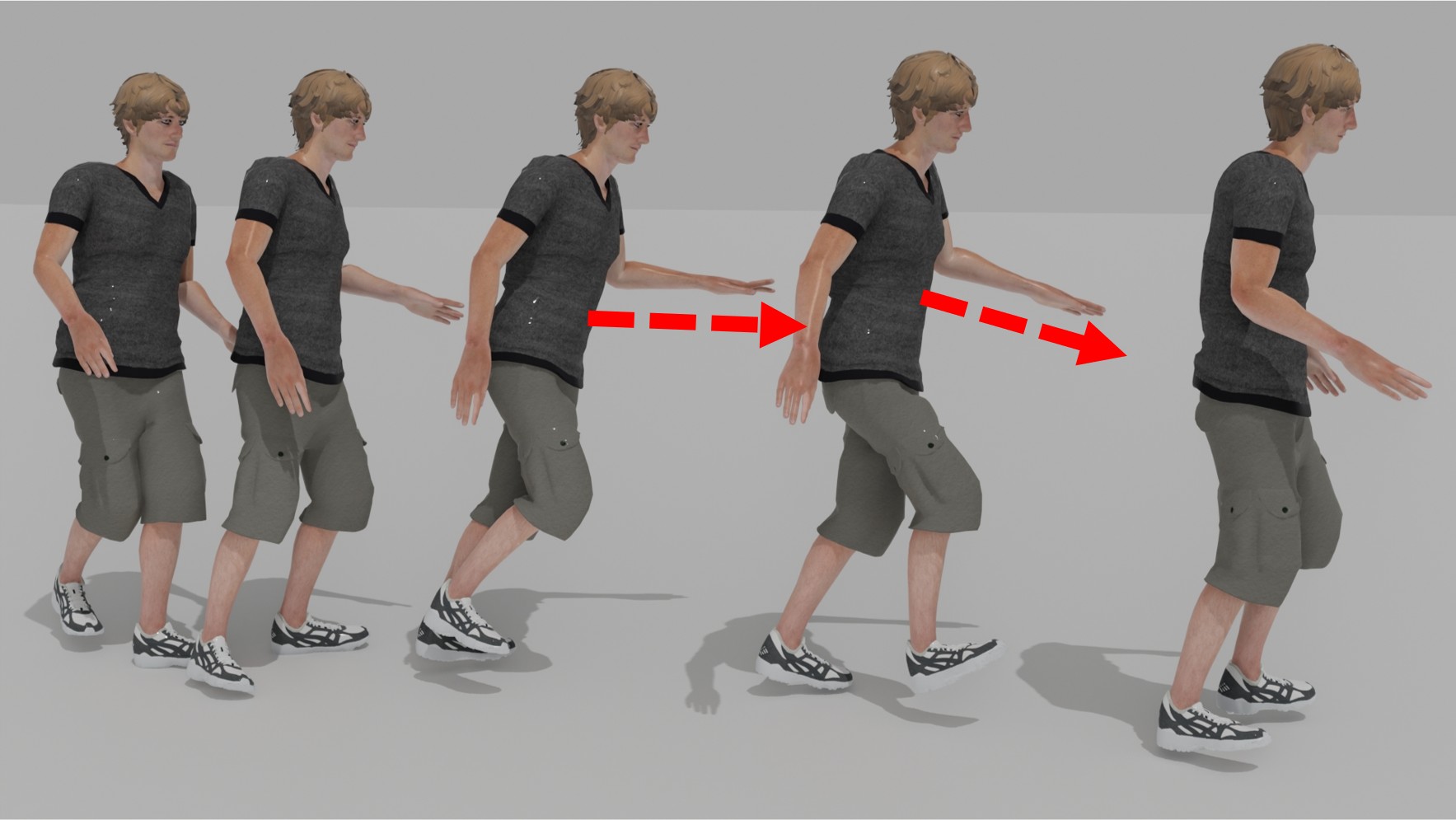}}  \\ 
    \midrule
    MotionPuzzle \cite{jang2022motion} & 
        \parbox[c]{3cm}{\centering  \includegraphics[width=3cm]{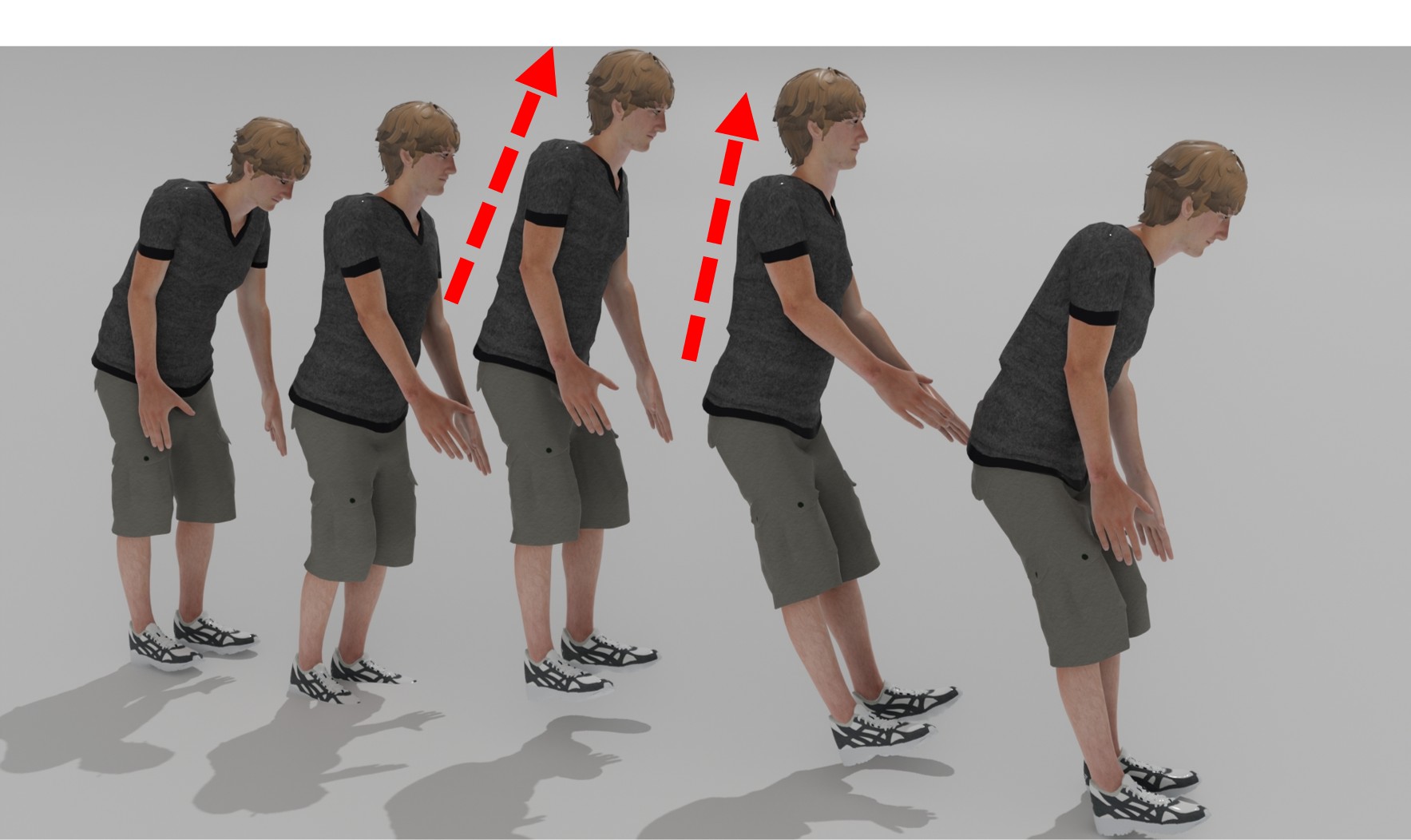}}  &
        \parbox[c]{3cm}{\centering  \includegraphics[width=3cm]{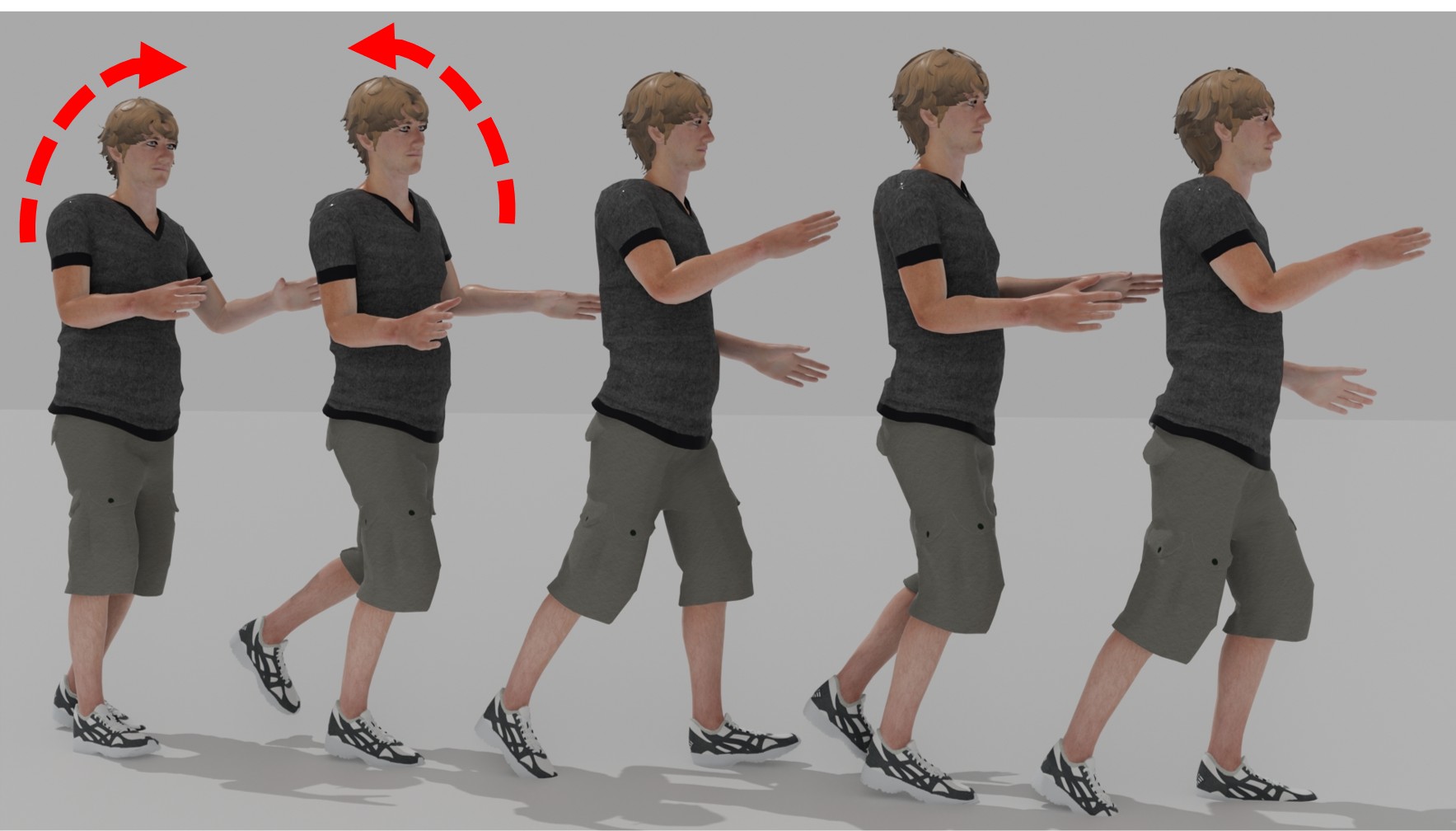} }  &
        \parbox[c]{3cm}{\centering  \includegraphics[width=3cm]{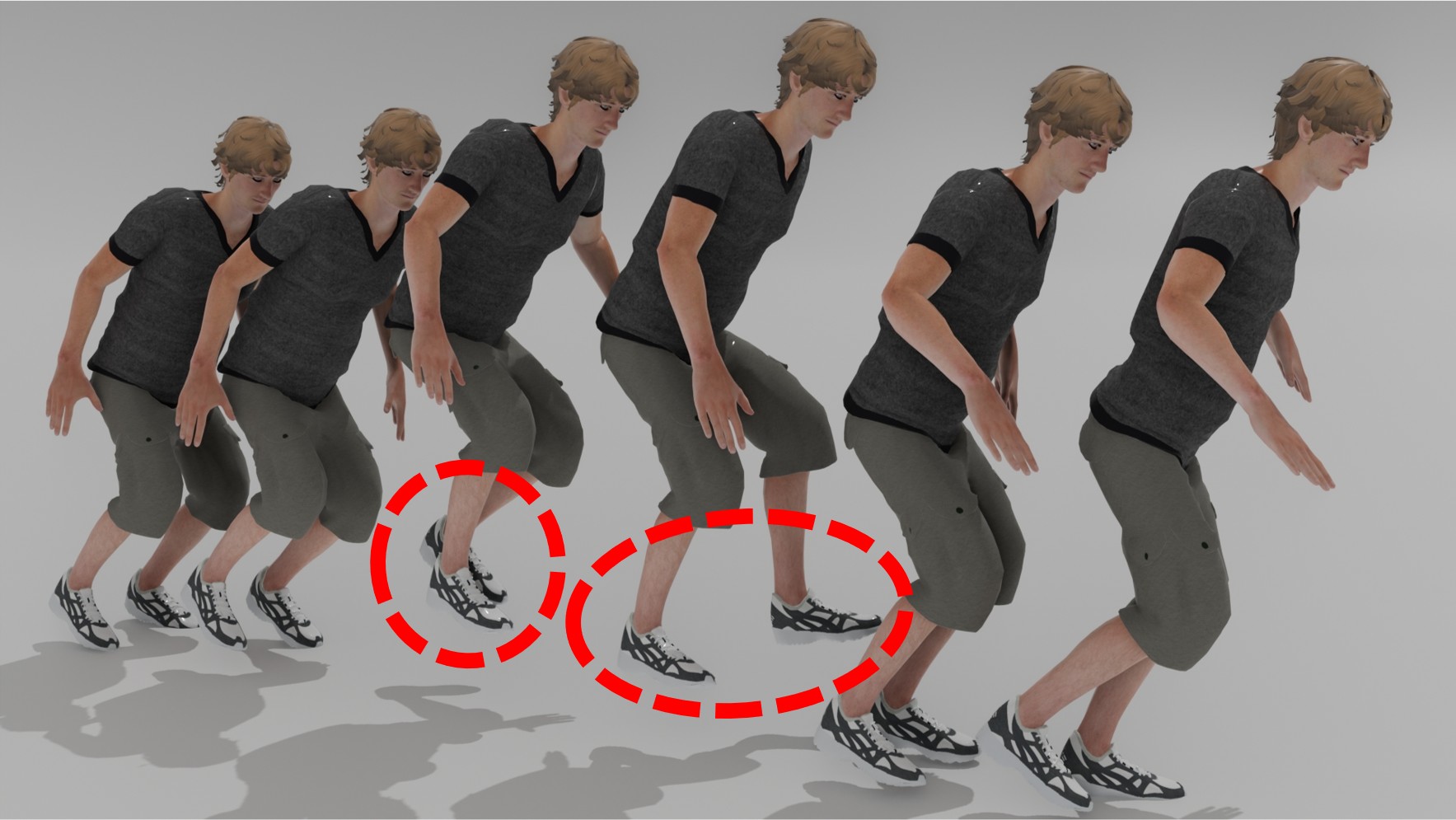} } & 
        \parbox[c]{3cm}{\centering  \includegraphics[width=3cm]{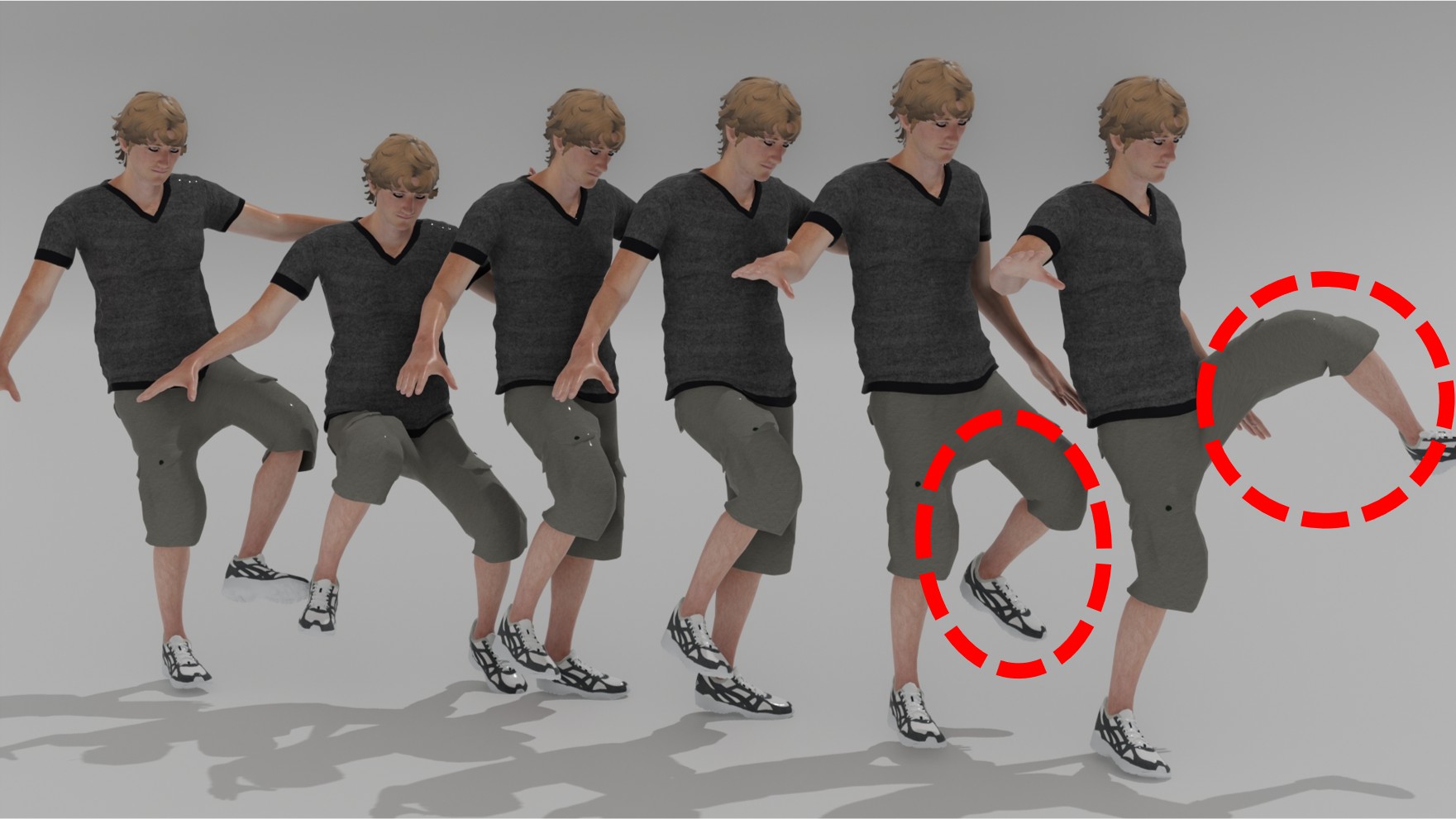 }}     \\
    \midrule
    Park et al. \cite{park2021diverse} &  
        \parbox[c]{3cm}{\centering  \includegraphics[width=3cm]{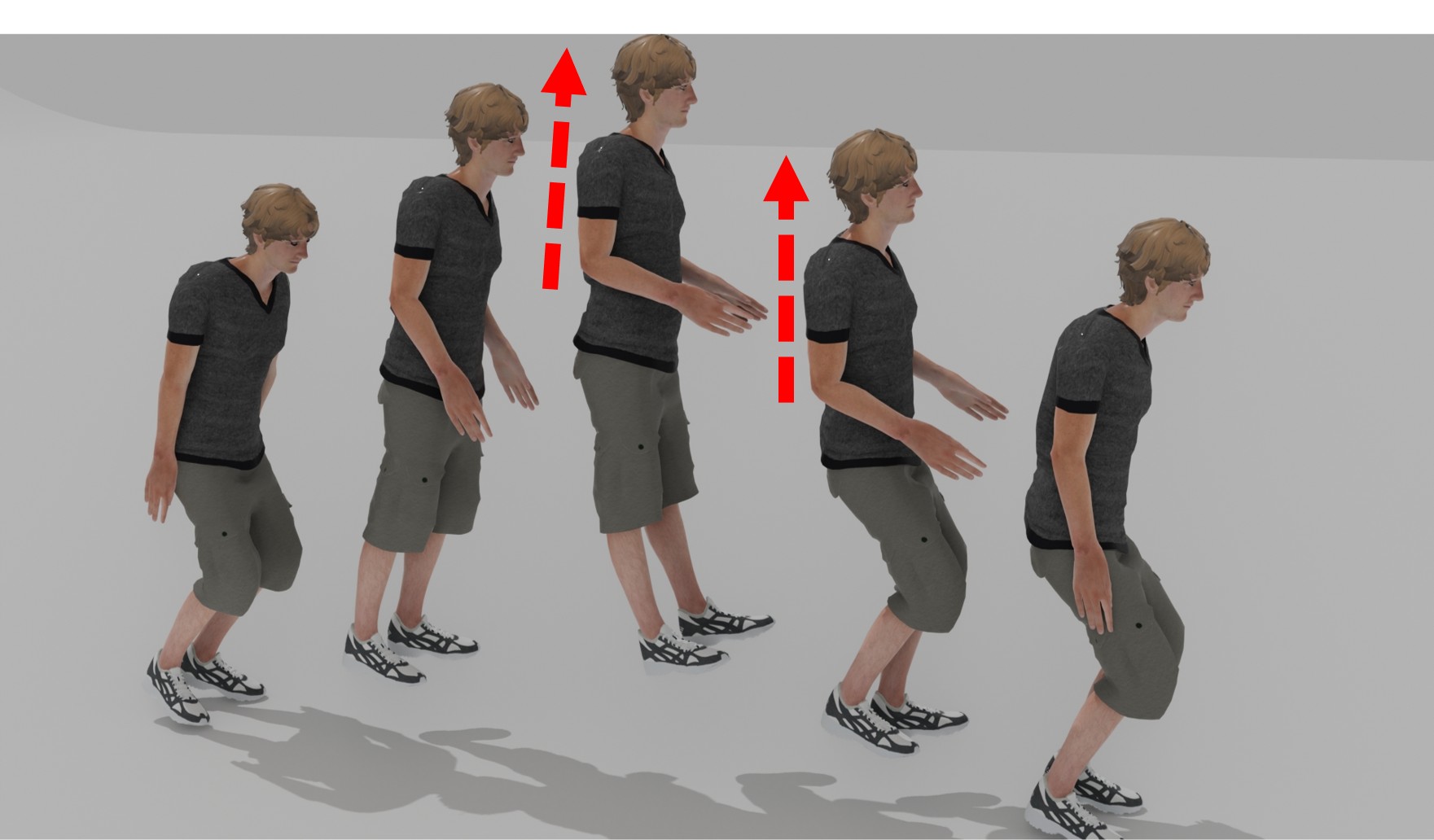} }  &
        \parbox[c]{3cm}{\centering  \includegraphics[width=3cm]{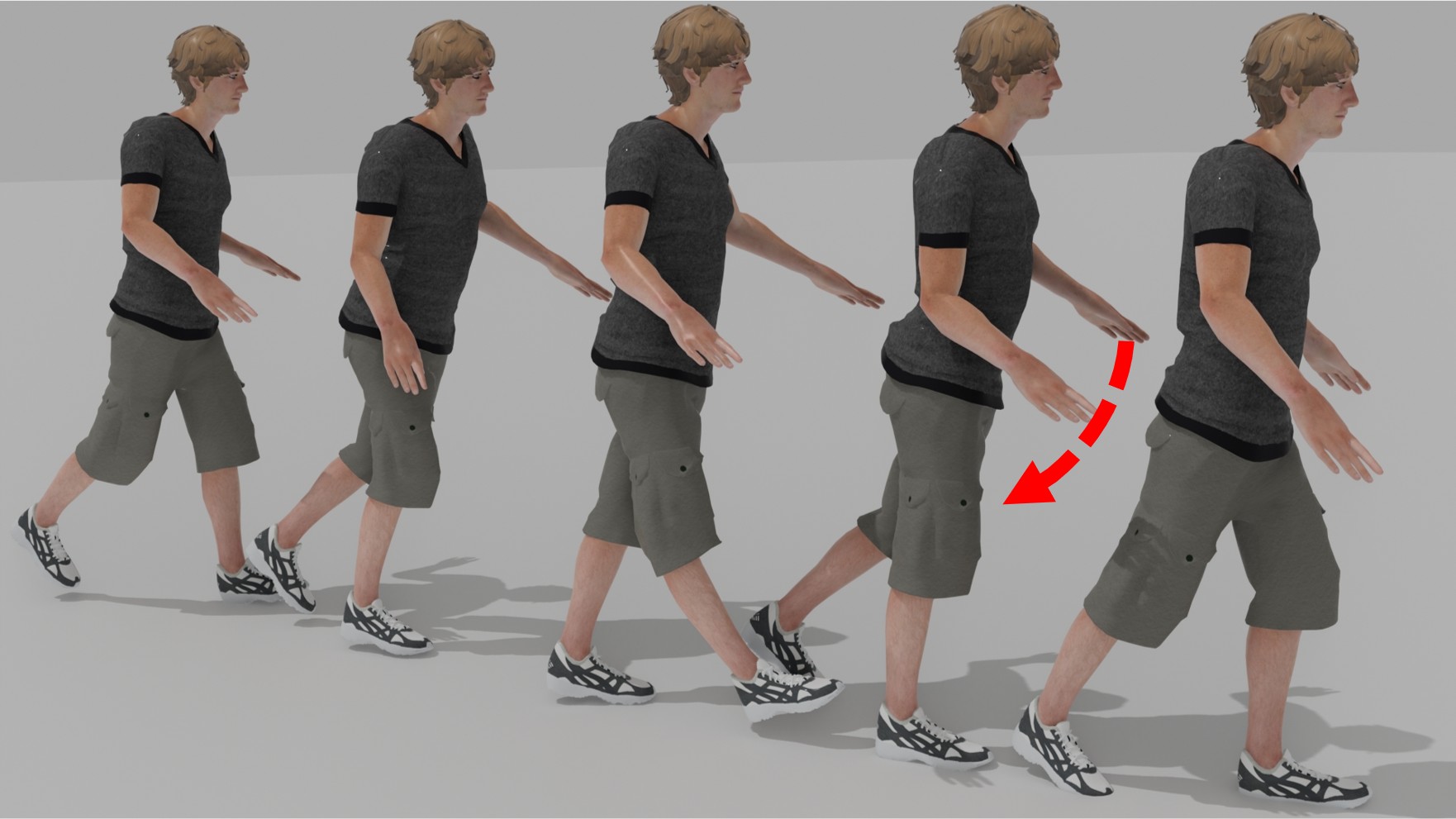}}  &
        \parbox[c]{3cm}{\centering  \includegraphics[width=3cm]{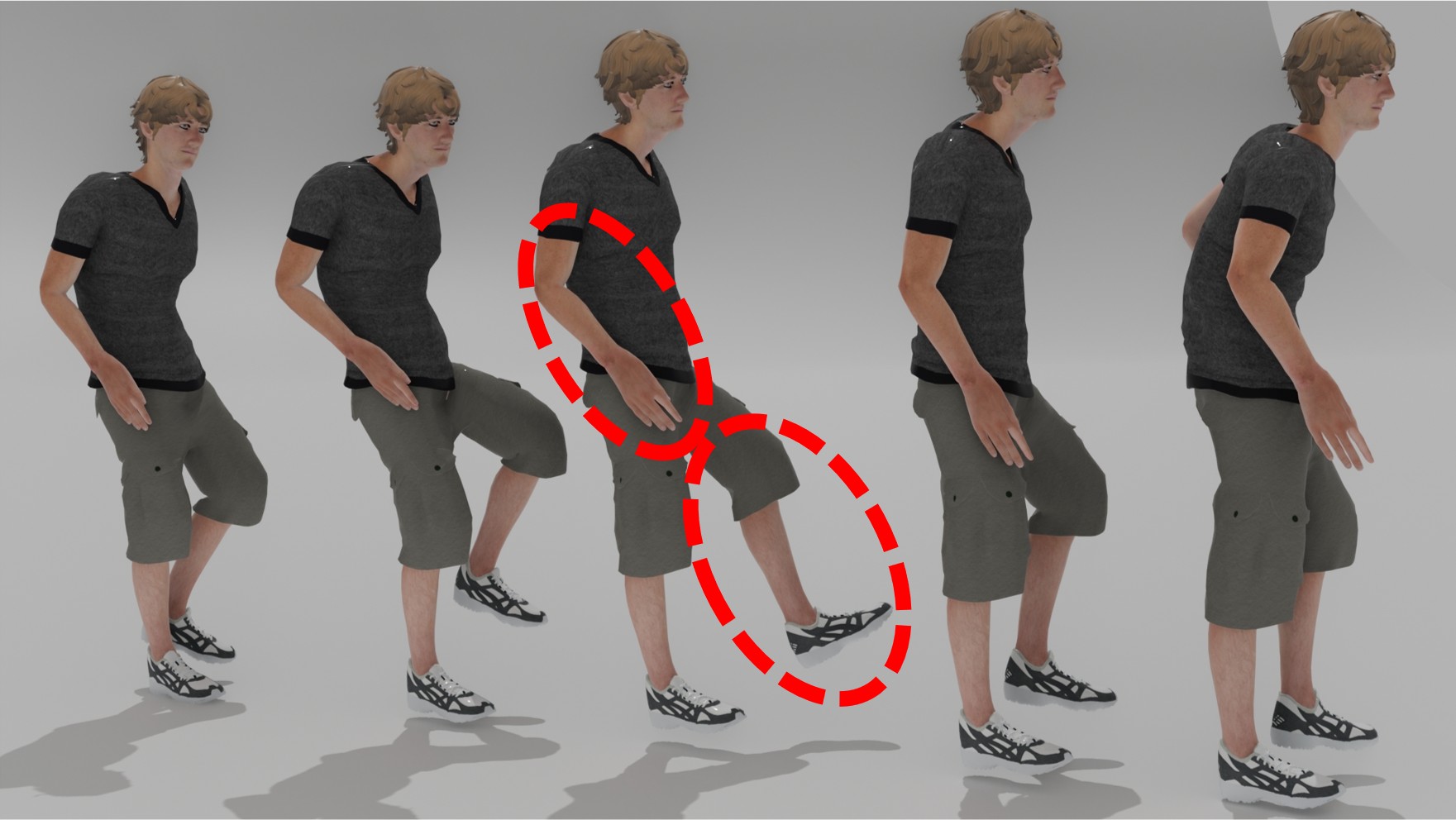} } &
        \parbox[c]{3cm}{\centering  \includegraphics[width=3cm]{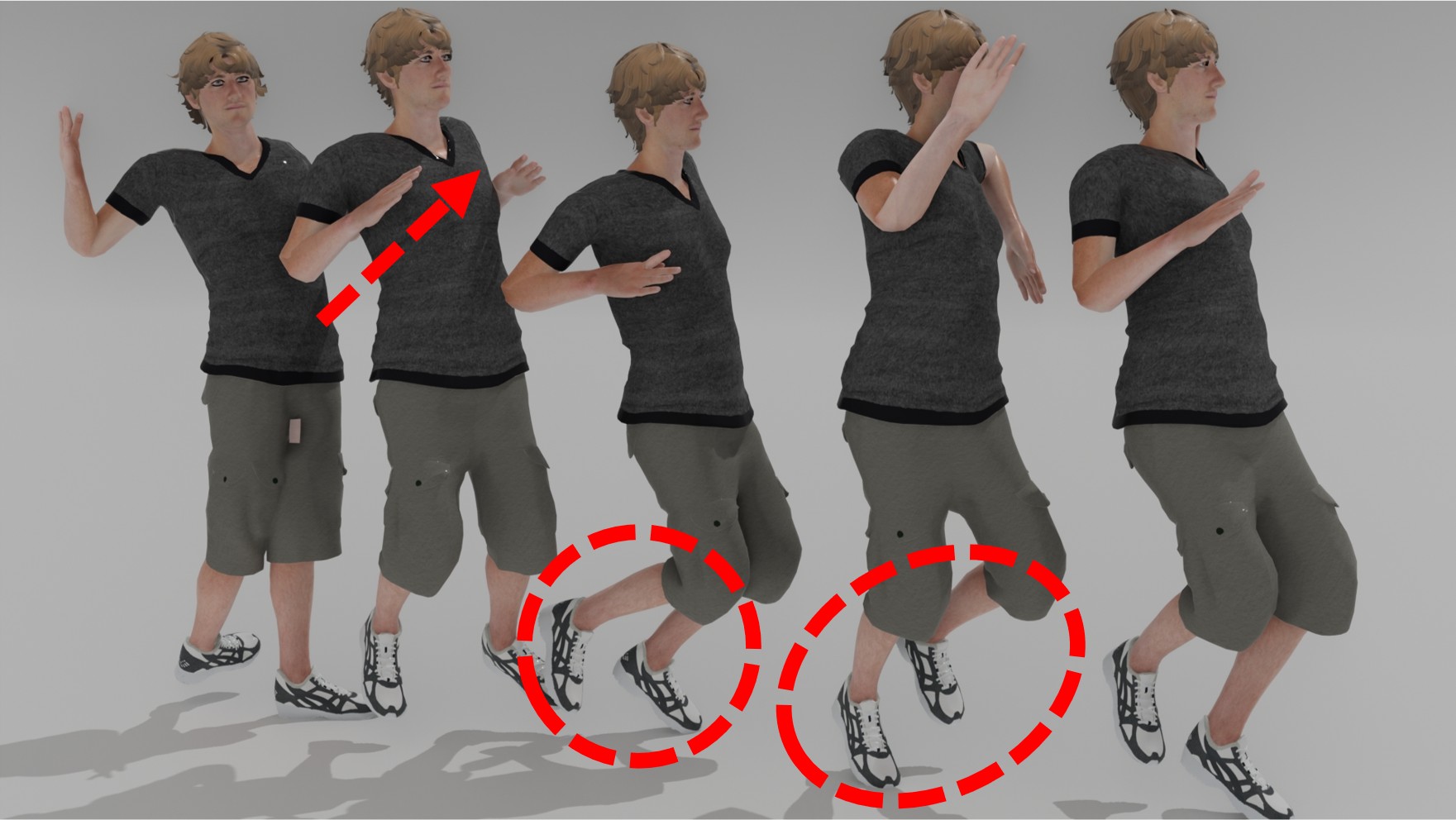}}   \\ 
    \bottomrule
  \end{tabular}
\end{table*}

\subsection{Evaluation Results}

\textbf{Quantitative Result.} In order to evaluation the content retention and style fidelity capability of our AStF, the the Fr$\Acute{e}$chet Inception Distance (FID)~\cite{heusel2017gans}, recognition accuracy and geodesic distance are utilized as quantitative measures. The detailed numerical quantitative results for BFA dataset ~\cite{aberman2020unpaired} and Xia dataset ~\cite{xia2015realtime} are presented in Table ~\ref{tab:quan}. As indicated in Table \ref{tab:quan}, our AStF achieves the lowest style FID and content FID, as well as the highest style and content accuracy, along with effective geodesic distance. This demonstrates that our AStF outperforms in terms of style fidelity and content retention. In contrast, other methods exhibit certain shortcomings in specific terms. Aberman et al. \cite{aberman2020unpaired} show the lowest accuracy and the highest FID, indicating failure in most test cases. MotionPuzzle \cite{jang2022motion} and Park et al. \cite{park2021diverse} perform well in style accuracy and style FID but has slight deficiencies in content retention, suggesting a weakness in preserving content. MoST \cite{kim2024most} excels in content accuracy and FID, and achieves better results in geodesic distance compared to our AStF, however, shows weaknesses in style fidelity, which means MoST tends to align more closely with content motion with insufficient style feature extraction. GenMoStyle ~\cite{guo2024generative} yields a satisfactory results on both style fidelity and content retention, however, there is still a noticeable gap compared to our method.

\textbf{Qualitative Result.} To further intuitively demonstrate the effectiveness of our AStF, we generated motions on the test sets of the Xia datset \cite{xia2015realtime} and BFA dataset \cite{aberman2020unpaired} with our trained AStF, the qualitative results are presented in Table \ref{tab:qualitative}. As Xia dataset ~\cite{xia2015realtime} is labeled by content, we present more qualitative results on Xia dataset. Since Aberman et al. failed in most cases, we have omitted its visualizations for the sake of clarity. For the purpose of clear demonstration, we applied red indifications in each figure in Table \ref{tab:qualitative}. Dashed circles and arrows indicate undesired content which may reflect to style characteristic in the content motion, while solid lines represent the desired content which need to be transferred from style motion to content motion. For instance, as in case (1) in Table \ref{tab:qualitative}, we take childlike jump as content motion and depressed walk as style motion. The action of arms-outstretched in the content motion is the content that we do not desired to retain, as it reflects a characteristics of childlike style. On the other hand, the head-down, hunched posture, which represents the style of depressed in the style motion, is what we aim to extract and transfer to the content motion. Additionally, the action of jump in content motion is also what we desired to retain. As presented in the experiment results of case (1), the generated motion of our AStF effectively transferred the depressed style while retaining the content jump. MoST~\cite{kim2024most} failed to discard the childlike characteristics in the content motion. MotionPuzzle \cite{jang2022motion} appeared not to adequately preserve the jump content. Park et al. ~\cite{park2021diverse} and Aberman et al. ~\cite{aberman2020unpaired} were insufficient in transferring depressed style to content motion. Overall, the qualitative experimental results demonstrate that our AStF effectively achieves style fidelity and content retention, successfully transferring style while preserving content. While  MoST ~\cite{kim2024most} and GenMoStyle ~\cite{guo2024generative} effectively preserving content, it fails to achieve sufficient style transfer, with generated motions tending to align more closely with content motion. MotionPuzzle ~\cite{jang2022motion}, Aberman et al. ~\cite{aberman2020unpaired} and Park et al. \cite{park2021diverse} show inadequate content preservation, producing awkward motions with indistinct styles.

\subsection{Ablation Study}

To validate the necessity of key components in our AStF, we conduct ablation experiments focusing on five critical elements: (1) proposed MCR Loss, (2) Simple SDM conducted in Statistics Encoder, (3) high-order statistics in SDM. (4) style-align loss, (5) hyper-parameters in our loss function. To thoroughly demonstrate the impact of these elements on content retention and style fidelity of our AStF, all experiments are conducted on the Xia dataset ~\cite{xia2015realtime}, since it has motions with labeled styles and contents. Table \ref{tab:ablation} presents the results of ablation experiments. Overall, through the aforementioned ablation experiments, we observed the significant impact of each module on enhancing the model's style fidelity and content retention. 
    
\textbf{The Impact of proposed MCR Loss.} The MCR Loss contains style-style MCR Loss and style-generation MCR Loss. We systematically evaluate their contributions through three ablation configurations: without style-style MCR loss, without style-generation MCR loss, and without the entire MCR loss. Our ablation experiments indicated that, both style-style and style-generation MCR Losses independently improved style and content FID and accuracy. The full model, compared to the variant without the entire MCR Loss, achieved a 8.7\% improvement in style accuracy and a 4.7\% improvement in content accuracy, highlighting the impact of MCR Loss on overall performance.

\textbf{The Impact of Simple SDM.} We tested the scenario of using a learnable style token instead of Simple SDM in our Statistics Encoder. The ablation experiments results indicate that using learnable style token significantly reduces style accuracy and results in the highest style FID value. In contrast, our Simple SDM achieves a 22.0\% improvement in style accuracy and a 0.046 enhancement in style FID.

\textbf{The Impact of high-order statistics.} Furthermore, as our AStF involves two additional high-order statistics, skewness and kurtosis, we assessed their importance in our SDM and Simple SDM by conducting three ablation experiments: removing skewness, removing kurtosis, and removing the both. The results indicate that using only mean and variance is insufficient. Adding skewness and kurtosis individually provides slight improvements, while incorporating both statistics results in a 7.2\% increase in style accuracy and a 4.6\% increase in content accuracy compared to without skewness and kurtosis.

\textbf{The Impact of proposed style-align loss.} To test the critical role of our proposed style-align loss function during generator training, we conducted an ablation experiment without it. The ablation experiment results show a significant decrease in style accuracy compared to full model, while content accuracy remained effective and geodesic distance reached the lowest value among all ablation experiments. This observation indicate that the generated motion tends to align more closely with content motion. However, in most cases, style features were insufficiently extracted from the style motion. In contrast, our style-align loss function resulted in a 29.3\% improvement in style accuracy.

\textbf{The Impact of different weights for losses.} As mentioned in Section \ref{subsec:loss}, MCR loss and style-align loss are novel losses specifically proposed in this work, while others are directly adopted in previous works, we only conducted ablation experiments on the the weight of MCR loss and style-align loss, which are indicated by $\lambda_\text{MCR}$ and $\lambda_\text{a}$. As indicated in Table \ref{tab:ablation}, while the weight of $\lambda_\text{MCR}$ increased, the performance of our AStF decreased in both content retention and style fidelity. As for style-align loss, Increasing $\lambda_\text{a}$ results in a decrease in style FID. However, it also causes an increase in content FID, indicating that an increased weight of style-align loss can lead to deficiencies in content retention.

\section{Conclusion}
In this work, we introduce AStF, a novel framework that effectively transfers style features from style motion to content motion. Our approach proposes using skewness and kurtosis along with mean and variance, to measure style features. We developed the SDM and HOS-Attn for effective extraction and integration of statistics features. Additionally, the MCR discriminator enhances style consistency through feature alignment. Expriments demonstrate our AStF outperforms existing methods in terms of style fidelity and content retention. However, our method lacks physical constraints on the joints and capturing subtle style features. In future work, we plan to expand our approach to enhance its capability in subtle styles and physical constraints.

\begin{acks}
Our work is supported in part by the Joint Fund of Ministry of Education of China (8091B022149, 8091B02072404), National Natural Science Foundation of China (62132016, 62302372), and Natural Science Basic Research Program of Shaanxi (2020JC-23).

\end{acks}

\bibliographystyle{ACM-Reference-Format}
\balance
\bibliography{sample-base}


\end{document}